\documentclass[journal]{IEEEtran}
\usepackage{blindtext}
\usepackage{amsmath, psfrag} 
\usepackage{amssymb, relsize}  
\usepackage{textcomp}
\usepackage{graphicx}
\usepackage[export]{adjustbox}
\usepackage{subcaption}
\usepackage{tikz}
\usepackage{pgfplots}
\usepackage{setspace}
\usepackage{multirow}
\usepackage[linesnumbered,ruled]{algorithm2e}
\usepackage{epstopdf,caption}
\newcommand{\subparagraph}{}
\usepackage{titlesec}
\titlespacing*{\section}{0pt}{0.6\baselineskip}{0.5\baselineskip}
\titlespacing*{\subsection}{0pt}{0.5\baselineskip}{0.4\baselineskip}
\titlespacing*{\subsubsection}{0pt}{0.4\baselineskip}{0.3\baselineskip}
\usetikzlibrary{positioning}
\usetikzlibrary{shapes.geometric, arrows}
\tikzstyle{startstop} = [rectangle, rounded corners, minimum width=1cm, minimum height=0.5cm,text centered, draw=black, fill=red!30, node distance = 1cm]
\tikzstyle{startstop1} = [rectangle, rounded corners, minimum width=1cm, minimum height=0.5cm,text centered, draw=black, fill=red!30, node distance = 1cm]
\tikzstyle{startstop2} = [rectangle,dashed, rounded corners, minimum width=1cm, minimum height=0.5cm,text centered, draw=black, fill=blue!30, node distance = 0.5cm]
\tikzstyle{arrow} = [thick,->,>=stealth]
\tikzstyle{arrow1} = [thick,<-,>=stealth]
\usepackage{booktabs}
\ifCLASSINFOpdf
 
\else
  
\fi

\begin{document}
\author{Neema Davis, Gaurav Raina, Krishna Jagannathan \thanks{N. Davis, G. Raina and K. Jagannathan are with the Department
of Electrical Engineering, Indian Institute of Technology Madras, Chennai 600 036, India. 
E-mail: \{ee14d212, gaurav, krishnaj\}@ee.iitm.ac.in}}
\title{Taxi demand forecasting: A HEDGE based tessellation strategy for improved accuracy}

\maketitle
\begin{abstract}
A key problem in location-based modeling and forecasting lies in identifying suitable spatial and temporal resolutions. In particular, judicious spatial partitioning can play a significant role in enhancing the performance of location-based forecasting models. In this work, we investigate two widely used tessellation strategies for partitioning city space, in the context of real-time taxi demand forecasting. Our study compares (i) Geohash tessellation, and (ii) Voronoi tessellation, using two distinct taxi demand datasets, over multiple time scales. For the purpose of comparison, we employ classical time-series tools to model the spatio-temporal demand. 
Our study finds that the performance of each tessellation strategy is highly dependent on the city geography, spatial distribution of the data, and the time of the day, and that neither strategy is found to perform optimally across the forecast horizon. 
We propose a hybrid tessellation algorithm that picks the best tessellation strategy at each instant, based on their performance in the recent past. 
Our hybrid algorithm is a non-stationary variant of the well-known HEDGE algorithm for choosing the best advice from multiple experts.
We show that the hybrid tessellation strategy performs consistently better than either of the two strategies across the data sets considered, at multiple time scales, and with different performance metrics.
We achieve an average accuracy of above 80\% per $\text{km}^2$ for both datasets considered at 60 minute aggregation levels.
\end{abstract}


\IEEEpeerreviewmaketitle

\section{Introduction} \label{intro}

Mobile application based e-hailing taxi services are gaining huge popularity across the world due to the advancement in GPS based smart phone technologies. These services can be used to supplement the use of public transit and other traditional modes of transportation. A key challenge faced by these fast growing taxi services is the demand - supply mismatch problem. During peak hours, the demand for taxis surpass the available supply, creating unmet demand. During off-peak hours, the scenario reverses and the vacant taxis cruise for longer periods to find passengers. These issues lead to dynamic surge pricing, along with reduced customer satisfaction and low driver utilization. Therefore, it is crucial to devise accurate location-specific demand forecasting algorithms, so as to gain prior knowledge of under-supplied and over-supplied areas. This knowledge can be used to mitigate the demand-supply imbalance by re-routing vacant taxi drivers, to compensate for the unmet demand. One of the key steps towards devising efficient location-based forecasting algorithms is the selection of proper spatial resolution for forecasting. The spatial resolution should not be too high, $i.e.,$ each cell should contain sufficient demand density for prediction to be effective. At the same time, the resolution should not be too low so that the driver has to cruise a large distance before finding a passenger. Hence, a carefully planned tessellation strategy is a key step in any location-based modeling exercise. 

\subsection{Related works}
Several attempts have already been made to study the supply-demand levels and imbalances in taxi services \cite{shao2015estimating,kamga2013hailing}. Identification and modeling of passenger hot spots for rerouting taxi drivers is also a widely researched area. Various methodologies have been proposed to this end, with Auto Regressive Integrated Moving Average model (ARIMA) and its variants \cite{li2012prediction, moreira2013predicting}, Exponential Weighted Average models \cite{zhang2016framework}, Nearest Neighbour clustering \cite{lee2008analysis,dong2017novel}, Neural Networks (NN)  \cite{xu2017real} being some of the commonly used modeling techniques. Irrespective of the modeling technique used, the preliminary step in modeling a location based entity such as passenger demand is to spatially partition the space. In the transportation literature, two common tessellation strategies are used. Spatial aggregations are either performed using grids, where the space is partitioned into square or rectangular grids of fixed area \cite{li2012prediction,xu2017real,phithakkitnukoon2010taxi}, or using polygons, where the space is partitioned into regular or irregular polygons of variable area \cite{ding2015dissecting,shen2017discovering,zhang2016learning}. It is common practice in the transportation literature to consider either one of these tessellation styles for spatial partitioning, without motivating the choice of the tessellation technique. 

In our previous work \cite{davis2016multi}, we tessellated the city of Bengaluru, India, into fixed-sized partitions known as geohashes.  We observed that for those regions with low demand density, fixed sized partitions resulted in data scarcity, which led to low model accuracy. To improve the accuracy, we explored the spatial correlation between the neighbouring geohashes to enhance the performance of the models. The performance limitation due to the chosen tessellation strategy motivated us to conduct a comparison study of various tessellation strategies. In this scenario, a few significant questions arise: (i) How can one decide the tessellation scheme to be used?, (ii) How sensitive is the performance of the models to the tessellation strategy?, (iii) Can we arrive at a tessellation strategy that works for a broad range of datasets? We aim to address these questions through this paper. To the best of our knowledge, an extensive study of the tessellation techniques and their effects on the model performance have not been conducted in the past. In this work, we explore the relationships between tessellation strategies, demand densities and city geographies. This is one of the features that distinguishes our work from the existing literature. 

For comparing fixed and variable sized tessellation styles, we choose Geohash tessellation, and Centroidal Voronoi tessellation with K-Means respectively. Intriguingly, each tessellation strategy has been shown to outperform the other \cite{outersterp2016partitioning, ravina2017voronoi}. In \cite{outersterp2016partitioning}, the authors claimed that Geohash technique works better at partitioning data than Voronoi technique for their data set, while in \cite{ravina2017voronoi}, authors preferred Voronoi over Geohash for forecasting supply of drivers in the top localities. In this work, we favour a partition based clustering technique such as K-Means \cite{macqueen1967some} over other clustering techniques. The reader is referred to \cite{fahad2014survey} and \cite{xu2015comprehensive} for a comprehensive survey of various clustering algorithms. In the survey paper \cite{xu2015comprehensive}, K-Means is listed as a potential clustering algorithm for large scale data, due to its low time complexity, and high computing efficiency. K-Means has a linear memory and time complexity, which is ideal for our very large data sets. It has also been shown that K-Means performs reasonably well in comparison with other clustering techniques; for example, DBSCAN \cite{chakraborty2014performance} and Hierarchical clustering \cite{kaushik2014comparative}, among others. In addition, K-Means has been widely used in the literature to generate Centroidal Voronoi polygons \cite{burns2009centroidal}. Of the widely used modeling techniques, Exponential Smoothing and ARIMA models are known for their simplicity in implementation and high computational speed \cite{moreira2015improving}. These techniques have also been observed to perform satisfactorily on comparison with other conventional methods such as Bayesian networks, SVR, and Artifical Neural Networks (ANN) \cite{chen2012retrieval,lippi2013short}. 
Authors in \cite{li2012prediction} proposed an improved ARIMA based prediction method to forecast the spatial-temporal variation of passengers in the hot spots with a prediction error of 5.8\%. Using a time varying Poisson process and ARIMA model, the work in \cite{moreira2013predicting} obtained an accuracy of 76\% for passenger demand on taxi services. Clustering along with Exponential Weighted Moving Average models were used to predict passenger demand hot spots with a 79.6\% hit ratio in \cite{zhang2016framework}. The subway ridership demand was analysed in \cite{ding2017using} using a combined ARIMA-GARCH model with a maximum error of 7\%. This promising performance of regression and smoothing based models in the context of passenger demand modeling, along with high computational speeds, make them ideal choices for our comparison study. 

Thus, we aim to perform an extensive city wide spatio-temporal analysis and compare the two tessellation techniques for two independent datasets -- an e-hailing mobile application based demand data set in Bengaluru, India and a  street hailing based taxi demand data set in New York, USA. Next, in addition to comparing the two strategies, we also propose a hybrid tessellation strategy by combining the two tessellation strategies based on their past performances, that works for any data set. No similar effort has been made in the transportation literature to combine tessellation strategies, and this feature also sets our work apart from the existing works. 

\subsection{Our contributions}
The demand data points are segregated into clusters using K-Means clustering algorithm. The obtained cluster centroids are used to partition the city into geohashes and Voronoi cells. In the past, road intersections \cite{ding2015dissecting,zhang2016learning} and bus stops \cite{liu2017intelligent} are employed to act as tessellation centers. In contrast, we use K-Means cluster centroids to act as the tessellation centers. Different time-series modeling techniques are applied to the Geohash aggregated data and Voronoi aggregated data to compare the two techniques. The data is aggregated and analyzed at two different time-scales; at 15 minutes to facilitate response to real-time decisions and, at 60 minutes to observe the high-level patterns. While comparing the two tessellation strategies, we observe that performances of the strategies do not remain consistent. Performances of the strategies are found to vary with different data sets. They also showed time dependent variations within each data set. In order to deal with this apparent non-stationarity, we need a combining algorithm that can pick the best tessellation strategy at every time instant. 

The method of using advises from multiple experts\footnote{In our case, the two ``experts" refer to the two tessellation strategies.} was first introduced in \cite{vovk1990aggregating}, and later generalized in \cite{freund1995desicion} to arrive at the well-known HEDGE algorithm. Specifically, by adopting a multiplicative update of weight parameter, the authors of \cite{freund1995desicion} were able to produce algorithms performing almost as well as the best expert in the pool. The authors in \cite{raj2017aggregating} extend the HEDGE algorithm to include non-stationary experts, by introducing a discounting factor. We modify the discounted variant of HEDGE to arrive at a hybrid tessellation strategy for enhanced taxi demand forecasts.  

The key contributions and findings of this paper can be summarized as follows:
\begin{enumerate}
    \item We conduct a comparative study of Voronoi and Geohash tessellation strategies for spatial demand partitioning.
     \item We highlight the dependence of the tessellation strategy on time of the day, and on the properties of the data set.
     \item  We find that models based on Voronoi tessellation have a superior performance compared to models based on Geohash tessellation for demand scarce cells. On the other hand, Geohash tessellation performs better than Voronoi tessellation in demand dense cells.
     \item We develop a hybrid tessellation strategy using a HEDGE based combining algorithm. We find that our hybrid strategy always performs at least as good as the best strategy out of the list of experts.
     \item Specifically, our algorithm picks the best tessellation strategy for each instant in the forecasting horizon, across various temporal and spatial resolutions.
\end{enumerate}
At a broader level, this work demonstrates the potential of improving location-based forecasts by efficiently partitioning large-scale temporal and spatial data for taxi hailing services. The rest of the paper is organized as follows: Section \ref{problemsetting} defines the problem, followed by a brief overview of the data, and the tessellation strategies in Section \ref{tessellation}. The time-series modelling procedure is explained, along with preliminary results in Section \ref{modelling}. The proposed model combining algorithm and the results are provided in Section \ref{hedgealgo}, and we conclude our work in Section \ref{conclusions}. 

\section{Problem setting} \label{problemsetting}
Let $y_t$ be the number of taxi bookings ($i.e.,$ demand) observed at time $t$ in a particular region $R$. In order to model $y$, we can represent the bookings as a time-sequence, where the sequence contains points aggregated over the space $R$ at time $t_1, \ldots, t_n$. The space $R$ can be fixed-sized or variable-sized. In our case, we denote them as geohash and Voronoi cell respectively. Geohash partitioning of a city is trivial, as it does not incorporate neighbour information or data volume. On the other hand, to perform Voronoi partitioning, we need to run a clustering algorithm to cluster nearby points together. In order to limit vacant taxi cruising, we aim to route drivers to regions of area no more than 1 $\text{km}^2$. The parameters in the clustering algorithm are set accordingly. 
We define the $density$ of a K-Means cluster centroid as the number of data points closer to that centroid than to any other centroid. We assume that a unique smallest distance K-Means centroid exists for each data point.

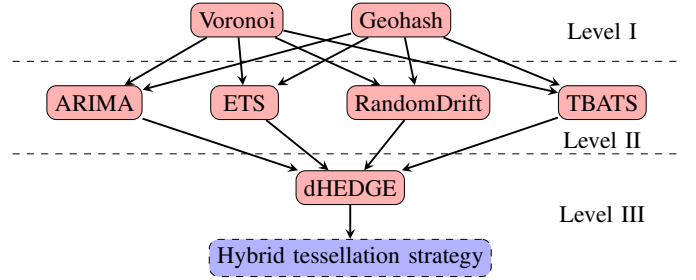
\begin{figure}
\centering
\scalebox{0.9}{
\begin{tikzpicture}[node distance = 0.75cm, auto]
    \node [startstop1] (vor) {Voronoi};
    \node [startstop1, right = of vor] (geo) {Geohash};
    \node [startstop, below left = of vor] (arima) {ARIMA};
    \node [startstop, right = of arima] (ets) {ETS};
    \node [startstop, right = of ets] (rd) {RandomDrift};
    \node [startstop, right = of rd] (tbats) {TBATS}(vor.south) -- coordinate (aux) (vor.south|-tbats.north);
    \node [above = of tbats,yshift = -0.2cm]{Level I};
     \draw[dashed] 
  ([xshift=-0.5cm]arima.north west|-aux) -- ([xshift=0.5cm]tbats.north east|-aux); 
     \path (arima) -- node (dh) [startstop,text=black,below=1cm] {dHEDGE} (tbats);
     \draw[dashed] ([xshift =-0.5cm,yshift =-0.5cm]arima.south west) -- ([xshift = 0.5cm,yshift = -0.5cm]tbats.south east);
     \node [below = of tbats, yshift= 0.7cm]{Level II};
     \node [below = of tbats, yshift= -0.4cm]{Level III};
     \node [startstop2,below = of dh](hybrid) {Hybrid tessellation strategy};
       \draw (vor)       edge[arrow] (arima)
                      edge[arrow] (ets)
                      edge[arrow] (rd)
                      edge[arrow] (tbats);
      \draw (geo)         edge[arrow] (arima)
                      edge[arrow] (ets)
                      edge[arrow] (rd)
                      edge[arrow] (tbats);
    \draw (dh)         edge[arrow1] (arima)
                      edge[arrow1] (ets)
                      edge[arrow1] (rd)
                      edge[arrow1] (tbats);
      \draw(dh)     edge[arrow] (hybrid);
                      \end{tikzpicture}}
\caption{Flow chart of the study.}
\label{flowchart}
\end{figure}

For an idle driver, we aim to provide him/her with the location of nearest suitable K-Means centroid. The demand aggregated from a Voronoi cell may vary in magnitufr from the demand aggregated from its corresponding geohash. Hence, for a standard comparison of the two techniques, we normalize the demand by the area of the partition to obtain demand per $\text{km}^2$. The area normalized demand $\text{D}_{norm}$ for $P^{th}$ geohash/Voronoi partition is computed as follows:
\begin{equation}
    \text{D}_{norm} = \frac{\text{aggregate(d) over $sp$ period}}{\text{area}(P)},
\end{equation}
where $d$ refers to the demand in P, $sp$ is the sampling period of 15 minutes or 60 minutes. This $\text{D}_{norm}$ is used to generate time-sequences for further analysis. The SMAPE and MASE \cite{hyndman2006another} are the error performance metrics that are considered in this work, and are defined in Section \ref{modelling}. See Figure \ref{flowchart} for a schematic representation of the work to be conducted. We are mainly interested in the Levels I and III of the flowchart, with nodes in Level II acting as tools for comparison. The data sets used for this study are mentioned below.

\subsection{Data set description}
The Bengaluru taxi demand data is acquired from a leading Indian e-hailing taxi service provider. The data contains GPS traces of taxi passengers booking a taxi by logging into the mobile app. The data is available for a period of two months, $\text{1}^{st}$ of January 2016 to $\text{29}^{th}$ of February 2016. The data set contains latitude-longitude coordinates of the logged in customer, along with his/her identification number, session duration and time stamp. The latitude and longitude coordinates of the city is 12.9716° N, 77.5946° E, with an area of approximately 740 $\text{km}^2$. 

The New York yellow taxi cab data set is publicly available at \cite{nyc}. The data set contains GPS traces of government-run street hailing Yellow taxis. This data set differs from our mobile app based data set, both in terms of data volume and city structure. The geographical structure of Bengaluru city is radial, while that of New York city is linear. We considered the period of January-February 2016, for analysis. We extract the pick up locations and time stamps from the data to form the demand set. The latitude and longitude coordinates of New York city is 40.7128° N, 74.0059° W, with an area of approximately 780 $\text{km}^2$. 

\begin{figure*}[htb]
\begin{subfigure}{0.375\textwidth}
\begin{tikzpicture}
\begin{axis}[enlargelimits=false, 
                 axis on top,
                 width=\textwidth,
                 xlabel={\footnotesize{Longitude}},
                 ylabel={\footnotesize{Latitude}},
                  ylabel near ticks,
                xlabel near ticks,
                 ytick={0.1,0.9},
                 yticklabels={\scriptsize{12.7},\scriptsize{13}},
                 xtick={76.1,76.5,76.9},
                 xticklabels={\scriptsize{77.3},\scriptsize{77.55},\scriptsize{77.8}},
                 yticklabel style={rotate=90},
                 scale = 0.8
                  ]
    \addplot graphics [ymin=0,ymax=1,xmin=76,xmax=77]{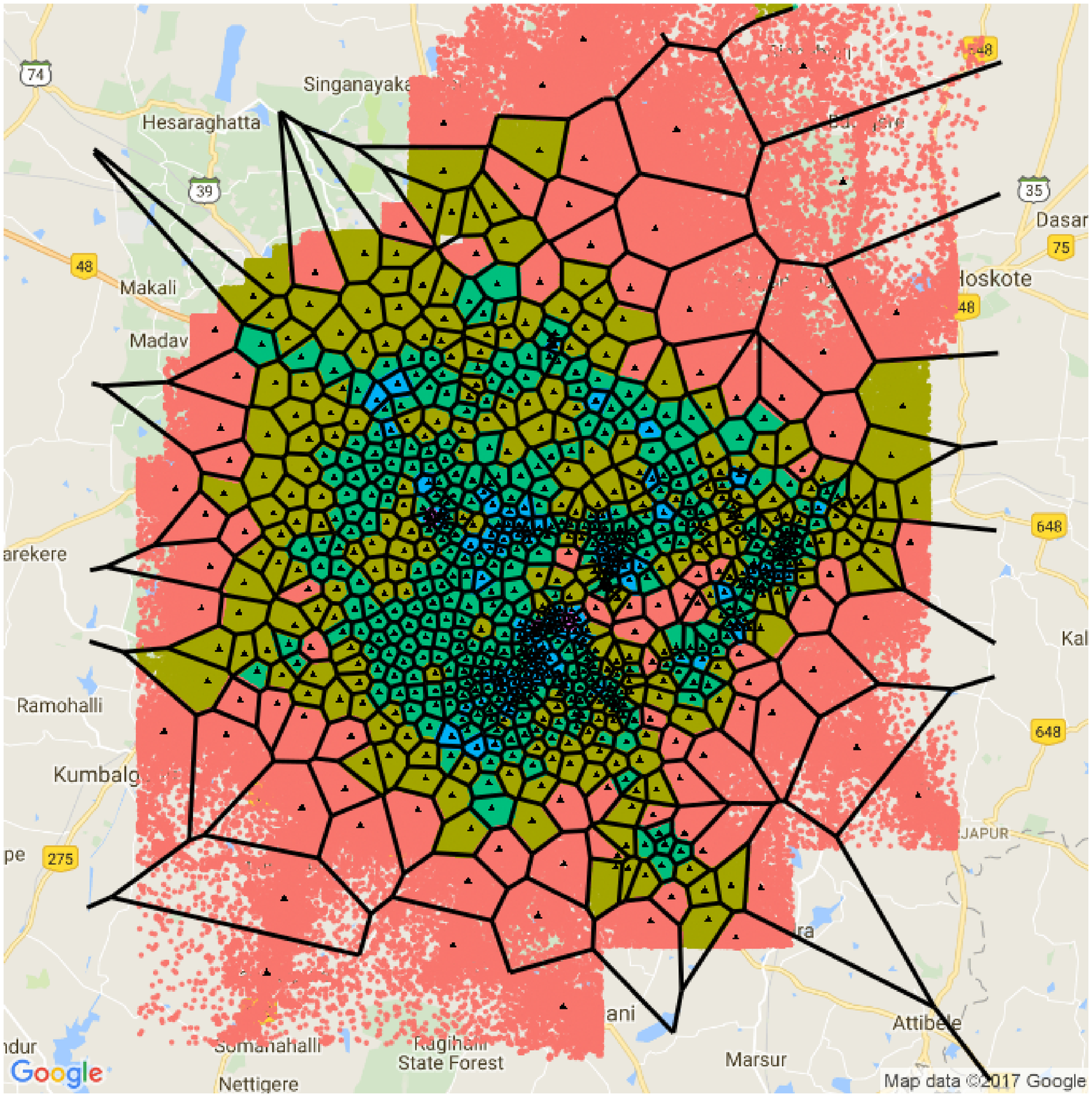};
    \end{axis}
 \end{tikzpicture}
 \vspace{-0.2cm}
  \caption{Voronoi tessellation}
\label{voronoi_hm}
\end{subfigure}%
\begin{subfigure}{0.375\textwidth}
\begin{tikzpicture}
\begin{axis}[enlargelimits=false, 
                 axis on top,
                 width=\textwidth,
                 xlabel={\footnotesize{Longitude}},
                 ylabel={\footnotesize{Latitude}},
                  ylabel near ticks,
                xlabel near ticks,
                 ytick={0.1,0.9},
                 yticklabels={\scriptsize{12.7},\scriptsize{13}},
                 xtick={76.1,76.5,76.9},
                 xticklabels={\scriptsize{77.3},\scriptsize{77.55},\scriptsize{77.8}},
                 yticklabel style={rotate=90},
                 scale = 0.8
                  ]
    \addplot graphics [ymin=0,ymax=1,xmin=76,xmax=77]{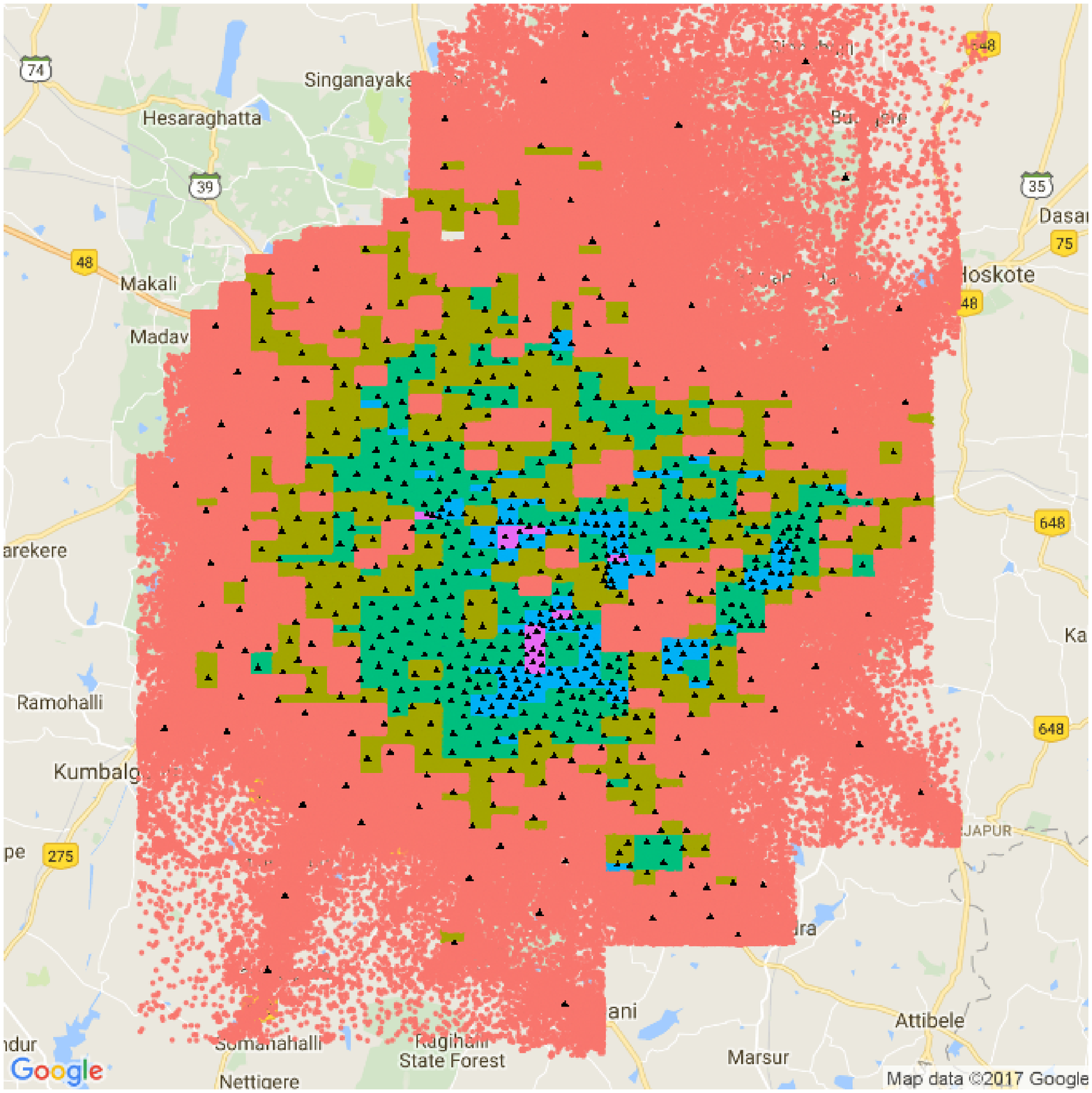};
    \end{axis}
\end{tikzpicture}
\vspace{-0.2cm}
 \caption{Geohash tessellation}
\label{geohash_hm}
\end{subfigure}%
\begin{subfigure}{0.2\textwidth}
\begin{picture}(100,100)
  \put(15,0)  {\includegraphics[trim={0 0 2cm 1cm},clip,scale=0.4]{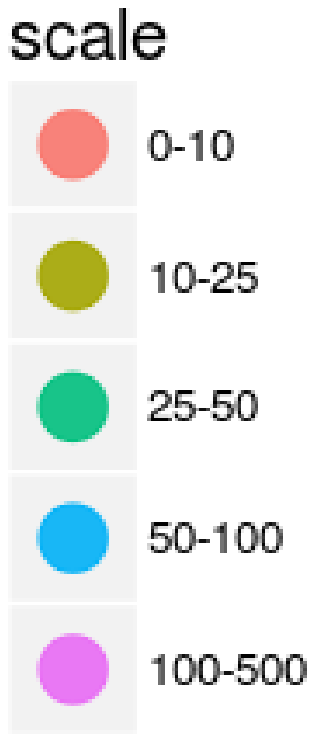}}
  \put(20,90) {SCALE}
  \put(35,70) {\footnotesize{0-7500}}
  \put(35,55) {\footnotesize{7500-17500}}
  \put(35,40) {\footnotesize{17500-35000}}
  \put(35,25) {\footnotesize{35000-70000}}
  \put(35,10) {\footnotesize{70000-230000}}
\end{picture}
\end{subfigure}%
\vspace{3mm}
\caption{Heatmaps obtained after partitioning Bengaluru city into Voronoi cells and 6-level geohashes.} \label{heatmaps} \vspace{5mm}
\end{figure*}

\begin{figure*}
\begin{subfigure}{0.375\textwidth}
\begin{tikzpicture}
\begin{axis}[enlargelimits=false, 
                 axis on top,
                 width=\textwidth,
                 xlabel={\footnotesize{Longitude}},
                 ylabel={\footnotesize{Latitude}},
                 ylabel near ticks,
                xlabel near ticks,
                  ytick={0.1,0.9},
                 yticklabels={\scriptsize{40.710},\scriptsize{40.711}},
                 xtick={76.1,76.5,76.9},
                 xticklabels={\scriptsize{-74.01},\scriptsize{-73.99},\scriptsize{-73.97}},
                 yticklabel style={rotate=90},
                 scale = 0.8
                  ]
    \addplot graphics [ymin=0,ymax=1,xmin=76,xmax=77]{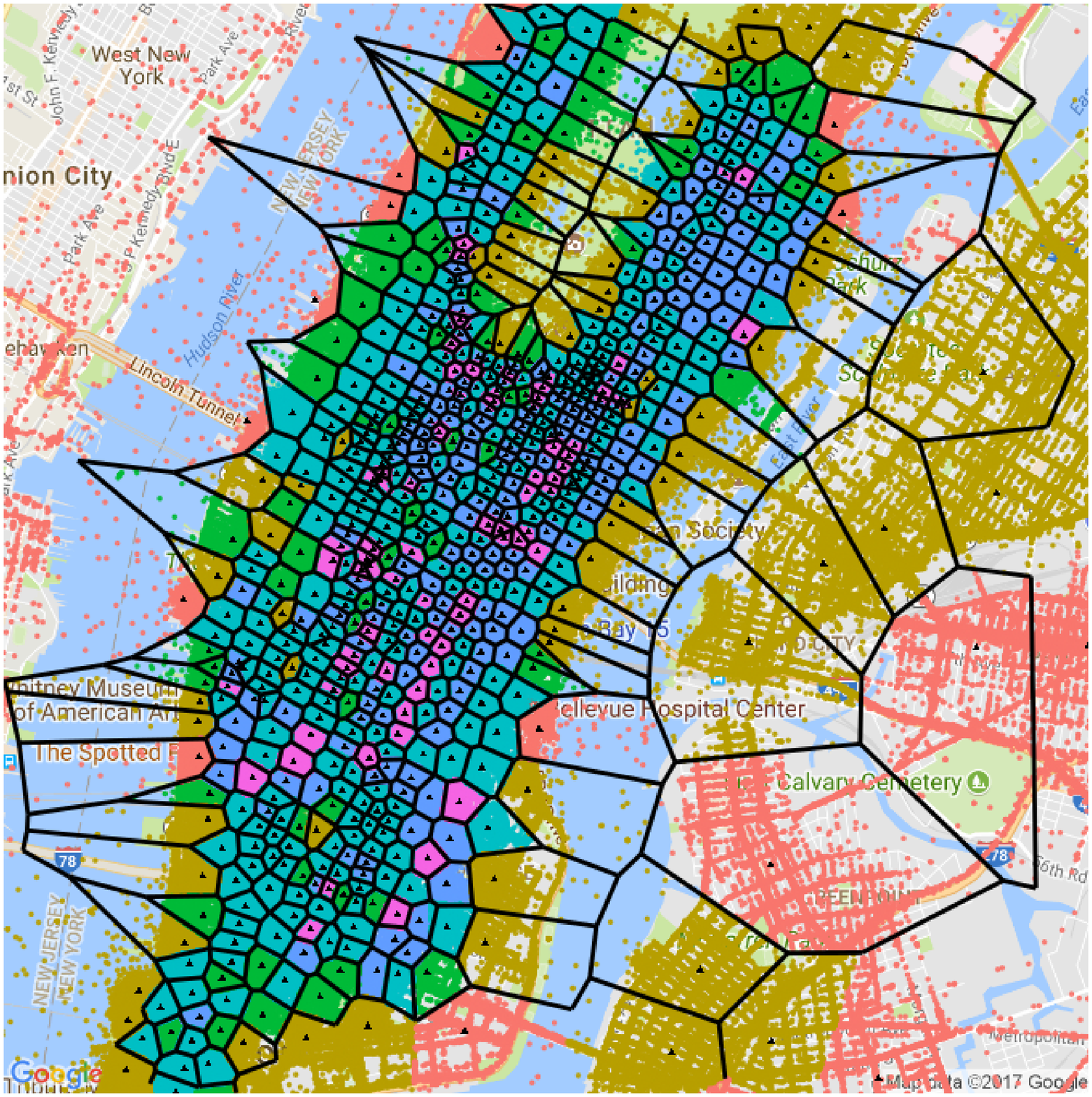};
    \end{axis}
 \end{tikzpicture}
  \vspace{-0.2cm}
  \caption{Voronoi tessellation}
\label{nyc_cluster}
\end{subfigure}%
\begin{subfigure}{0.375\textwidth}
\begin{tikzpicture}
\begin{axis}[enlargelimits=false, 
                 axis on top,
                 width=\textwidth,
                 xlabel={\footnotesize{Longitude}},
                 ylabel={\footnotesize{Latitude}},
                  ylabel near ticks,
                xlabel near ticks,
                 ytick={0.1,0.9},
                  yticklabels={\scriptsize{40.710},\scriptsize{40.711}},
                 xtick={76.1,76.5,76.9},
                 xticklabels={\scriptsize{-74.01},\scriptsize{-73.99},\scriptsize{-73.97}},
                 yticklabel style={rotate=90},
                 scale = 0.8
                  ]
    \addplot graphics [ymin=0,ymax=1,xmin=76,xmax=77]{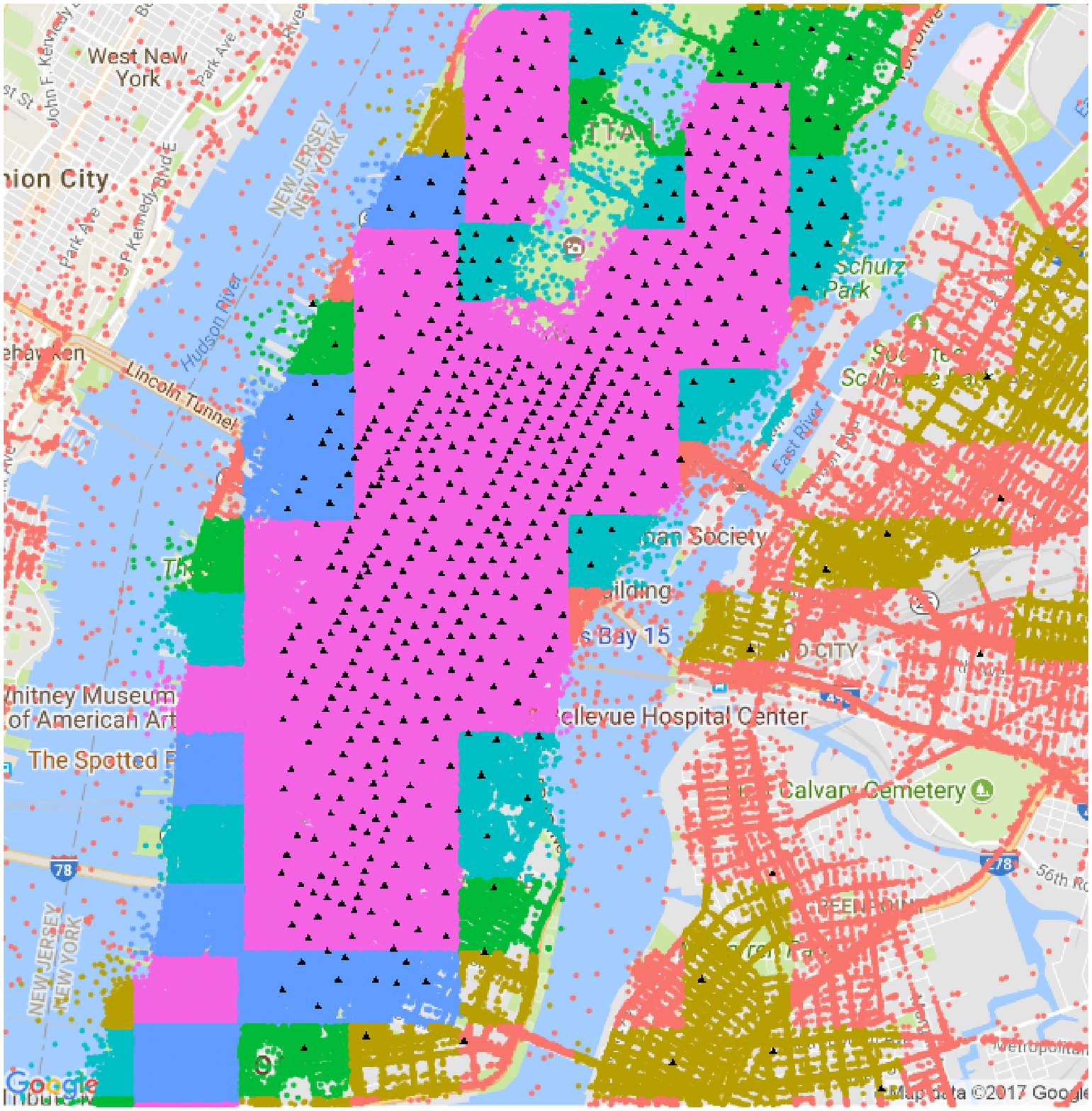};
    \end{axis}
\end{tikzpicture}
 \vspace{-0.2cm}
 \caption{Geohash tessellation}
\label{nyc_geohash}
\end{subfigure}%
\begin{subfigure}{0.2\textwidth}
\begin{picture}(100,100)
  \put(15,0)  {\includegraphics[trim={0 0 2.1cm 1cm},clip,scale = 0.4]{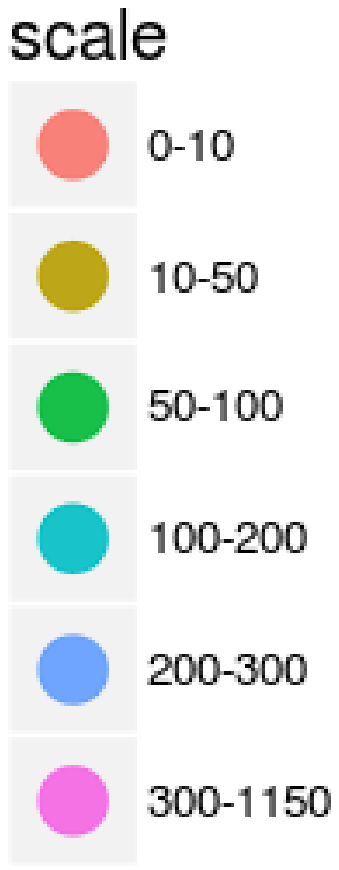}}
  \put(20,100) {SCALE}
  \put(35,82) {\footnotesize{0-7500}}
  \put(35,68) {\footnotesize{7500-35000}}
  \put(35,53) {\footnotesize{35000-70000}}
  \put(35,38) {\footnotesize{70000-140000}}
  \put(35,23) {\footnotesize{140000-220000}}
  \put(35,8) {\footnotesize{220000-250000}}
\end{picture}
\end{subfigure}%
\vspace{3mm}
\caption{Heatmaps obtained after partitioning New York city into Voronoi cells and 6-level geohashes.} \label{nyc_hm}
\end{figure*}

\section{Tessellation strategies} \label{tessellation}
As motivated in Section \ref{problemsetting}, we perform spatial partitioning of the city using two tessellation strategies for the purpose of modeling and comparison. First, we generate clusters of demand points for each city. Later, the centroids of these clusters aid in the generation of tessellation cells. 
\subsection{K-Means algorithm}
K-Means \cite{macqueen1967some} is a widely used unsupervised learning algorithm to classify a given data set through a certain number of clusters (assume $k$ clusters) fixed apriori. This algorithm aims at minimizing the squared error function given as:
\begin{equation}
    J = \sum_{j=1}^k \sum_{i=1}^n ||y_{(i)}^j - c_j||^2,
\end{equation}
where $||y_{(i)}^j - c_j||^2$ is the Euclidean distance between a data point $y_{(i)}^j$ and its center $c_j$, $n$ is the total number of data points. For ensuring that the average cluster area remains around 1 $\text{km}^2$, the number of centers $K$ is set to 740 and 780 for clustering Bengaluru and New York city respectively. For each data point, the algorithm calculates the distance from the data point to each cluster. If the data point is closest to its own cluster, leave it where it is, else, move it into the closest cluster. The algorithm stops when no data point is reassigned.
By sorting these clusters in descending order of \emph{density}, we can find locations that generate relatively higher demand (hot spots) compared to other locations.
\subsection{Voronoi tessellation}
Voronoi tessellation is a spatial partitioning method that divides space according to the nearest neighbour-rule. Specifically, each point, called a seed or site, is associated with the region ($i.e.,$ Voronoi cell) that is closer to it than to all other points in the space. A Voronoi tessellation is called centroidal when the generating site of each Voronoi cell is also its mean (center of mass). In our work, these sites are obtained from the K-Means algorithm. Based on the closeness of sites, this tessellation strategy produces polygon partitions of varying areas. For eg., the 740 demand centers result in 740 variable-sized quadrilateral partitions for Bengaluru city. We remark that the overall time complexity of Voronoi and K-Means algorithm is \emph{O(nlogn)}. Refer Figures \ref{voronoi_hm} and \ref{nyc_cluster} for heat maps generated from voronoi tessellations of Bengaluru and New York city. The partitioned cells are color-coded according to their sample size, for ease of representation on the color scale. 
\subsection{Geohash tessellation}
Geohash is a technique that hashes the latitude and longitude coordinates into a character string. It is an extension of grid based method, with a simple naming convention. Note that with Geohash (first G capitalized), we refer to the \emph{technique} of encoding a coordinate pair into a single string, while geohash (all small letters) refers to the string itself. Geohash can be visualized as a division of Earth into 32 planes, each of which can be divided again into 32 planes, and so on. We refer to these divisions as Geohash levels by defining level x as the division that results in geohashes of length x. A 6-level geohash spans a grid of area 1.2 km $\times$ 0.6 km, covering approximately 1 $\text{km}^2$, where a 5-level geohash spans an area of 4.9 km $\times$ 4.9 km, covering approximately 25 $\text{km}^2$. We choose 6-level grids over 5-level as it is more sensible to route drivers to a smaller area. In terms of time complexity, this algorithm is \emph{O(1)}.  Refer Figures \ref{geohash_hm} and \ref{nyc_geohash} for heat maps obtained from geohash tessellations of Bengaluru and New York city.
 \begin{figure*}
                \begin{subfigure}{0.25\textwidth}
                    \psfrag{area}{\hspace{-7mm} \raisebox{1.5mm}{\footnotesize{Area (in $\text{km}^2$)}}}
                	\psfrag{freq}{\hspace{-3mm}\raisebox{0mm}{\footnotesize{Frequency}}}
                	\psfrag{clu}{\hspace{0mm}\raisebox{0mm}{\footnotesize{Voronoi}}}
                	\psfrag{geo}{\hspace{0mm}\raisebox{0mm}{\footnotesize{Geohash}}}
                	\psfrag{0}{\hspace{0mm}\raisebox{0mm}{\footnotesize{0}}}
                	\psfrag{1}{\hspace{0mm}\raisebox{2mm}{\footnotesize{}}}
                	\psfrag{2}{\hspace{0mm}\raisebox{0mm}{\footnotesize{2}}}
                	\psfrag{3}{\hspace{0mm}\raisebox{2mm}{\footnotesize{}}}
                	\psfrag{4}{\hspace{0mm}\raisebox{0mm}{\footnotesize{4}}}
                	\psfrag{5}{\hspace{0mm}\raisebox{2mm}{\footnotesize{}}}
                	\psfrag{20}{\hspace{0mm}\raisebox{0mm}{\footnotesize{}}}
                	\psfrag{40}{\hspace{0mm}\raisebox{0mm}{\footnotesize{40}}}
                	\psfrag{60}{\hspace{0mm}\raisebox{0mm}{\footnotesize{}}}
                	\psfrag{80}{\hspace{0mm}\raisebox{0mm}{\footnotesize{80}}}
                	\psfrag{100}{\hspace{0mm}\raisebox{0mm}{\footnotesize{}}}
                    \includegraphics[scale = 0.4]{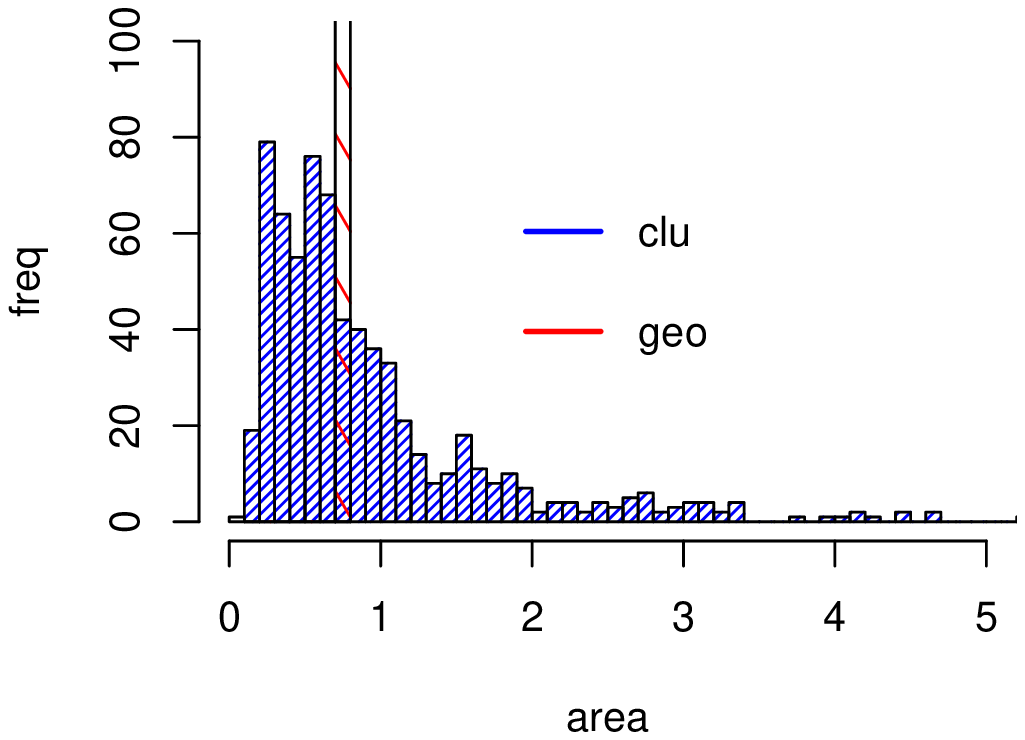}
                    \vspace{-0.75cm}
                    \caption{Cell area (Bengaluru)}
                    \label{bang_combined_area}
                \end{subfigure}%
                \begin{subfigure}{0.25\textwidth}
                    \psfrag{s}{\hspace{-7mm}\raisebox{1.5mm}{\footnotesize{Sample size}}}
                   	\psfrag{freq}{\hspace{-3mm}\raisebox{0mm}{\footnotesize{Frequency}}}
                   	\psfrag{clu}{\hspace{0mm}\raisebox{0mm}{\footnotesize{Voronoi}}}
                	\psfrag{geo}{\hspace{0mm}\raisebox{0mm}{\footnotesize{Geohash}}}
                  	\psfrag{0}{\hspace{0mm}\raisebox{0mm}{\footnotesize{0}}}
                	\psfrag{30000}{\hspace{0mm}\raisebox{0mm}{\footnotesize{30000}}}
                	\psfrag{60000}{\hspace{0mm}\raisebox{0mm}{\footnotesize{60000}}}
                	\psfrag{50}{\hspace{0mm}\raisebox{0mm}{\footnotesize{50}}}
                	\psfrag{100}{\hspace{0mm}\raisebox{0mm}{\footnotesize{100}}}
                	\psfrag{150}{\hspace{0mm}\raisebox{0mm}{\footnotesize{}}}
                    \includegraphics[scale = 0.4]{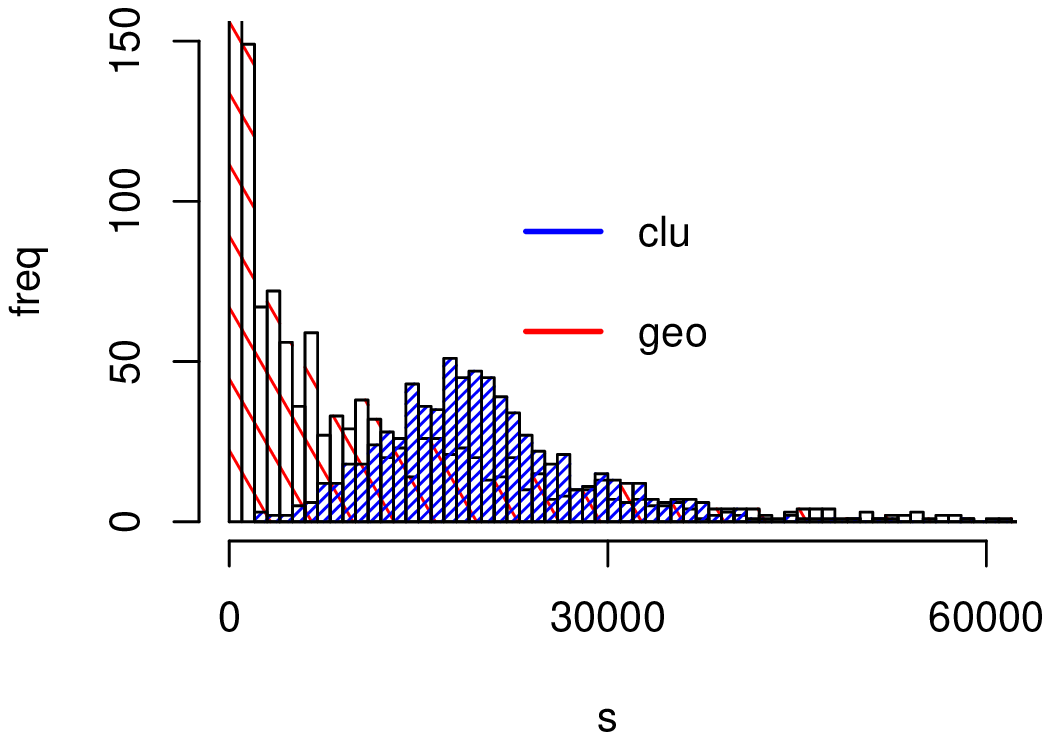}
                    \vspace{-0.75cm}
                    \caption{Cell volume (Bengaluru)}
                    \label{bang_combined_size}
                \end{subfigure}%
                    \begin{subfigure}{0.25\textwidth}
                    \psfrag{area}{\hspace{-7mm} \raisebox{1.5mm}{\footnotesize{Area (in $\text{km}^2$)}}}
                	\psfrag{freq}{\hspace{-3mm}\raisebox{0mm}{\footnotesize{Frequency}}}
                	\psfrag{clu}{\hspace{0mm}\raisebox{0mm}{\footnotesize{Voronoi}}}
                	\psfrag{geo}{\hspace{0mm}\raisebox{0mm}{\footnotesize{Geohash}}}
                	\psfrag{0.0}{\hspace{0mm}\raisebox{0mm}{\footnotesize{0}}}
                	\psfrag{0.5}{\hspace{0mm}\raisebox{0mm}{\footnotesize{}}}
                	\psfrag{1.0}{\hspace{0mm}\raisebox{0mm}{\footnotesize{1}}}
                	\psfrag{1.5}{\hspace{0mm}\raisebox{0mm}{\footnotesize{}}}
                	\psfrag{2.0}{\hspace{0mm}\raisebox{0mm}{\footnotesize{2}}}
                	\psfrag{0}{\hspace{0mm}\raisebox{0mm}{\footnotesize{0}}}
                	\psfrag{50}{\hspace{0mm}\raisebox{0mm}{\footnotesize{50}}}
                	\psfrag{100}{\hspace{0mm}\raisebox{0mm}{\footnotesize{100}}}
                	\psfrag{150}{\hspace{0mm}\raisebox{0mm}{\footnotesize{}}}
                    \includegraphics[scale = 0.4]{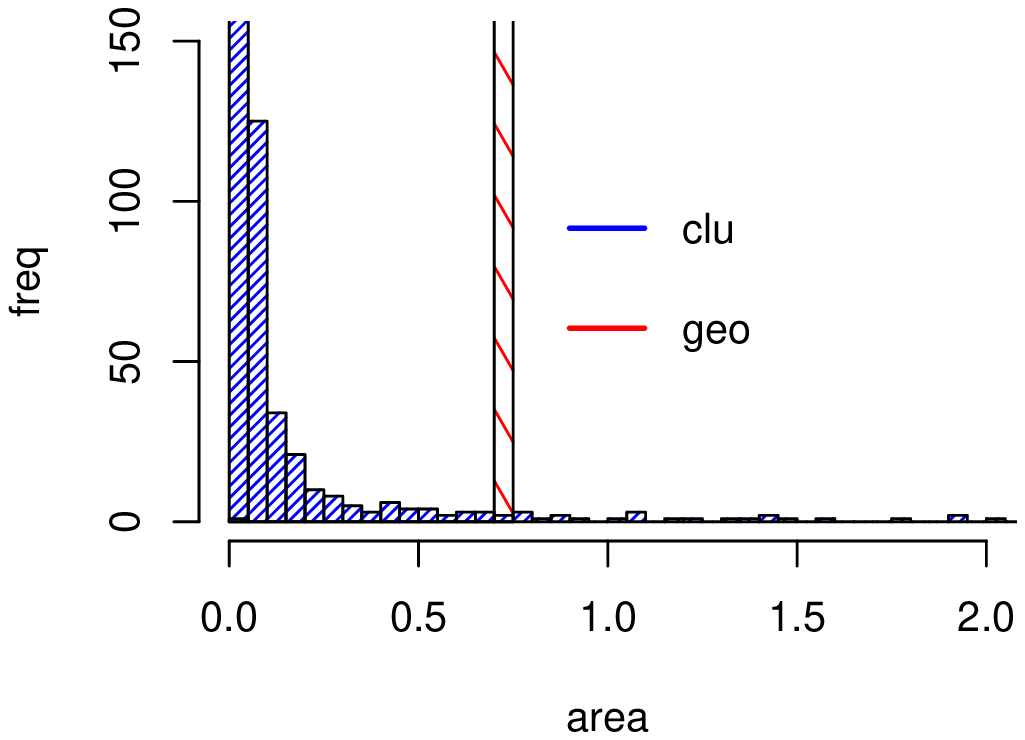}
                    \vspace{-0.75cm}
                    \caption{Cell area (New York)}
                    \label{nyc_combined_area}
                \end{subfigure}%
                \begin{subfigure}{0.25\textwidth}
                    \psfrag{s}{\hspace{-7mm}\raisebox{1.5mm}{\footnotesize{Sample size}}}
                   	\psfrag{freq}{\hspace{-3mm}\raisebox{0mm}{\footnotesize{Frequency}}}
                   	\psfrag{clu}{\hspace{0mm}\raisebox{0mm}{\footnotesize{Voronoi}}}
                	\psfrag{geo}{\hspace{0mm}\raisebox{0mm}{\footnotesize{Geohash}}}
                	\psfrag{0}{\hspace{0mm}\raisebox{0mm}{\footnotesize{0}}}
                	\psfrag{50}{\hspace{0mm}\raisebox{0mm}{\footnotesize{}}}
                	\psfrag{100}{\hspace{-2mm}\raisebox{0mm}{\footnotesize{100}}}
                	\psfrag{150}{\hspace{0mm}\raisebox{0mm}{\footnotesize{}}}
                	\psfrag{200}{\hspace{-2mm}\raisebox{0mm}{\footnotesize{200}}}
                  	\psfrag{0}{\hspace{0mm}\raisebox{0mm}{\footnotesize{0}}}
                	\psfrag{40000}{\hspace{0mm}\raisebox{0mm}{\footnotesize{40000}}}
                	\psfrag{80000}{\hspace{0mm}\raisebox{0mm}{\footnotesize{80000}}}
                    \includegraphics[scale = 0.4]{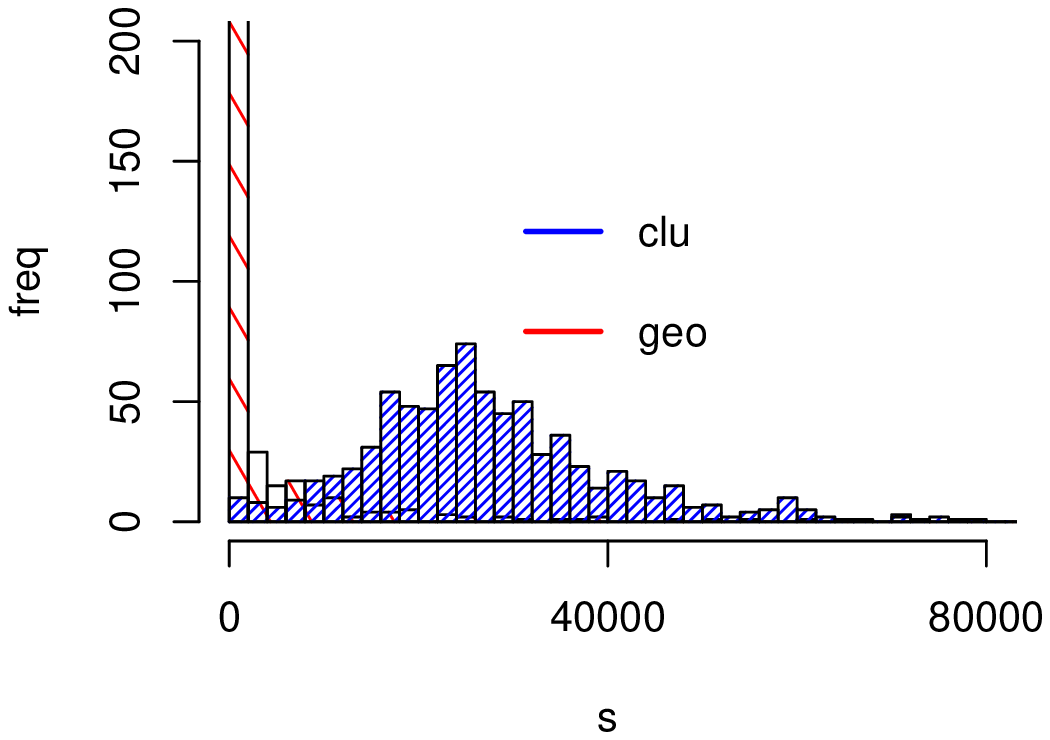}
                    \vspace{-0.75cm}
                    \caption{Cell volume (New York)}
                    \label{nyc_combined_size}
                \end{subfigure}%
                \vspace{5mm}
    \caption{Histogram to show the distribution of area per partition and samples per partition for the two tessellation techniques.}
    \label{bang_hist}
 \end{figure*}
\subsection{Observations}
\subsubsection{Bengaluru data set}
On referring Figure \ref{heatmaps} for the heat maps generating by the two tessellation techniques, we make the following inferences. Geohash is a region-oriented city map partition approach and therefore, results in some cells that are highly dense and in some cells that are highly sparse. On the other hand, Voronoi cells are more uniformly distributed in terms of density. Similar inferences can be made from the histogram plots (refer Figures \ref{bang_combined_area} and \ref{bang_combined_size}). The cell sample sizes follow a normal distribution for Voronoi cells in Figure \ref{bang_combined_size}. The tessellation technique tries to uniformly distribute samples among the partitions. This, in fact, increases the chance of finding a passenger in a cell as there are less ``demand scarce" cells. On the other hand, this process of uniformly distributing samples increases the partition size to above 1 $\text{km}^2$ for some Voronoi cells, as seen in Figure \ref{bang_combined_area}. As a result, the driver might have to traverse a larger distance to find a passenger. The size of the smallest Voronoi partition observed is 0.10 $\text{km}^2$ and 85\% of the partitions are below 2 $\text{km}^2$. The area of a geohash cell remains constant at 0.72 $\text{km}^2$. So when the driver is re-routed to a cell, if sufficient demand density exists in the cell, a vacant taxi driver has to spend less time searching in a geohash when compared to its corresponding Voronoi cell. At times, this advantage comes at the expense of masking the actual demand hot spots. In some locations, typically near the city center, there can be more than one hot spot in one $\text{km}^2$ area. Due to the inflexible structure of a geohash, these multiple hot spots are considered as a single hot spot. Note that if the service providers are more inclined towards re-allocating drivers to fixed sized locations, multiple hot spots in a geohash might not be an issue. With Voronoi cells, we have the added flexibility of sub-setting a geohash, into further cells. So from a hot spot identification view point, Voronoi tessellation seems to be a better strategy than Geohash tessellation for the Bengaluru data set.  

\subsubsection{New York data set}
Figure \ref{nyc_hm} is a zoomed-in heat map of borough of Manhattan.  We note that 92.5\% of the total generated demand originated from Manhattan over the period of study. Thus, the spatial data distribution is highly non-linear, which results in very closely packed Voronoi cells in the Manhattan borough. In fact, the smallest partition is of the size 0.08 $\text{km}^2$ and 75\% of the partitions are below 0.10 $km^2$, as observed in Figure \ref{nyc_combined_area}. Routing drivers to such a small area is infeasible and economically not viable for any service provider. The Geohash strategy, on the other hand, is not a density-dependent technique and hence, assigns geohashes uniformly. From Figure \ref{nyc_cluster}, we observe that there are inter-island tessellations, even though all the centroids are in mainland. This does not make sense from a driver re-routing view point. This problem does not arise in Figure \ref{nyc_geohash} and the 6-level geohashes are confined to either sides of the river. Here, Voronoi tessellation fails as a spatial tessellation strategy because of the non-uniform spatial data distribution, and the city geography, which is spread over multiple islands. If Voronoi tessellation is to perform better, extensive parameter tuning of the K-Means algorithm has to be conducted. The Voronoi tessellation has to be performed on each borough to avoid inter-island tessellations, which adds to further complexity. For this data set, the sheer simplicity of the Geohash tessellation makes it the winning strategy. 
\begin{figure*}
                \begin{subfigure}{0.33\textwidth}
                    \psfrag{f}{\hspace{-5mm} \raisebox{0mm}{\footnotesize{Frequency}}}
                	\psfrag{r}{\hspace{-9mm}\raisebox{2mm}{\footnotesize{Residual Magnitude}}}
                	\psfrag{-2}{\hspace{0mm}\raisebox{0mm}{\footnotesize{-2}}}
                	\psfrag{-1}{\hspace{0mm}\raisebox{0mm}{\footnotesize{-1}}}
                	\psfrag{0}{\hspace{0mm}\raisebox{0mm}{\footnotesize{0}}}
                	\psfrag{1}{\hspace{0mm}\raisebox{0mm}{\footnotesize{1}}}
                    \includegraphics[height = 2in]{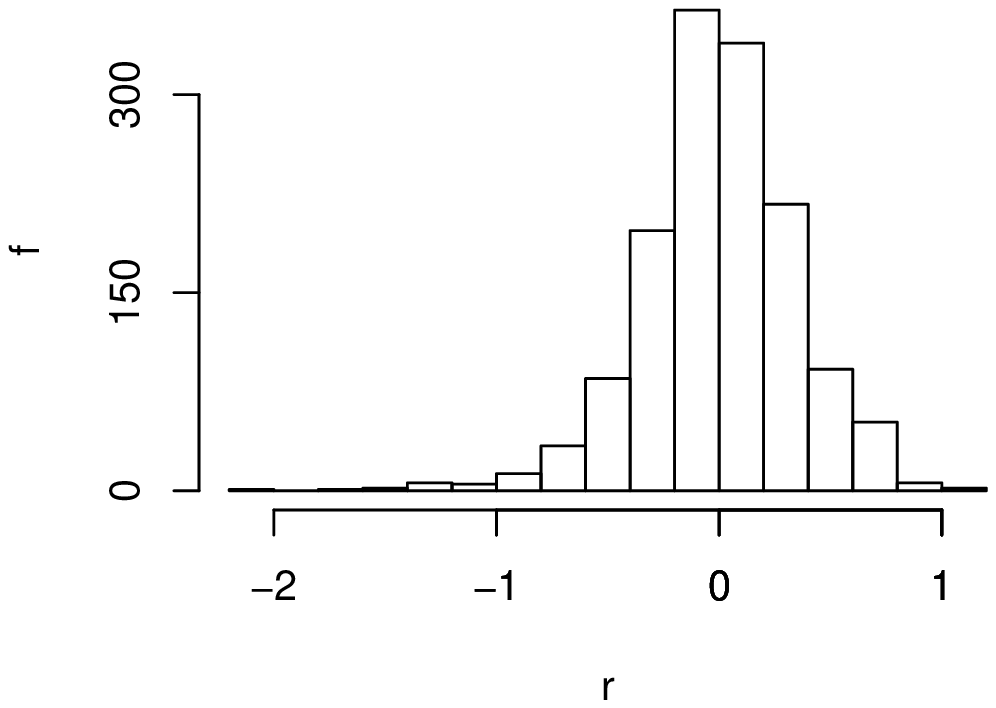}
                    \vspace{-0.5cm}
                    \caption{Sample histogram of residuals}
                \end{subfigure}%
                \begin{subfigure}{0.33\textwidth}
                    \psfrag{acf}{\hspace{-3mm} \raisebox{0mm}{\footnotesize{ACF}}}
                	\psfrag{l}{\hspace{-3mm}\raisebox{2mm}{\footnotesize{Lag}}}
                	\psfrag{0.0}{\hspace{0mm}\raisebox{-0mm}{\footnotesize{0}}}
                	\psfrag{0.5}{\hspace{0mm}\raisebox{0mm}{\footnotesize{}}}
                	\psfrag{1.0}{\hspace{0mm}\raisebox{-0mm}{\footnotesize{1}}}                	
                	\psfrag{1.5}{\hspace{0mm}\raisebox{0mm}{\footnotesize{}}}
                	\psfrag{2.0}{\hspace{0mm}\raisebox{-0mm}{\footnotesize{2}}}
                    \includegraphics[height = 2in]{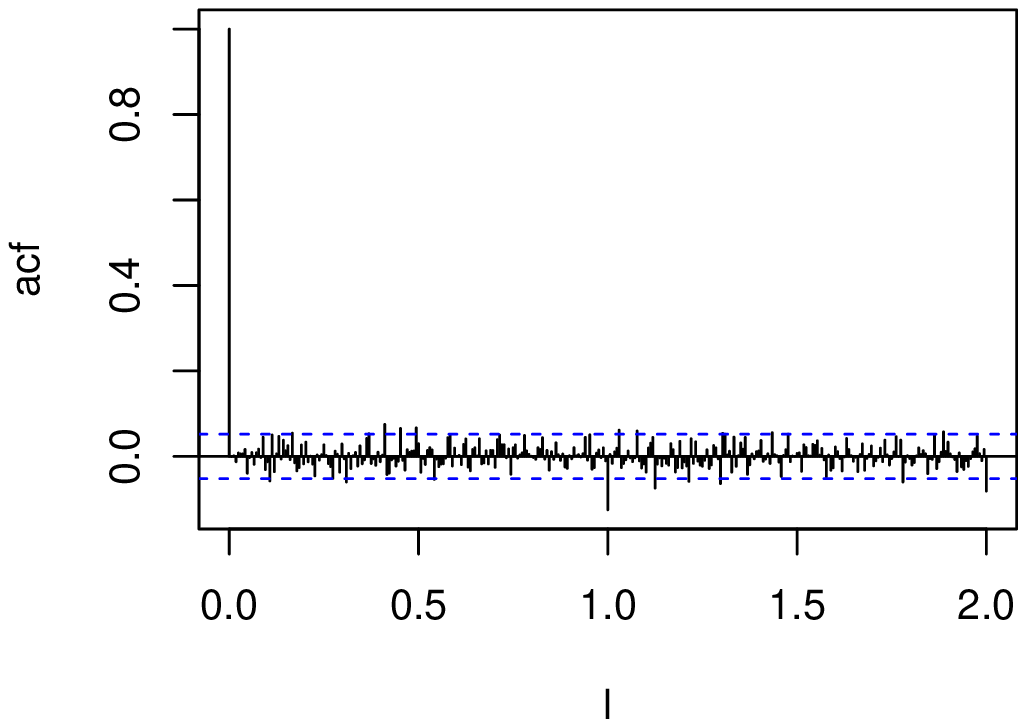}
                    \vspace{-0.5cm}
                    \caption{Significance levels}
                \end{subfigure}%
                    \begin{subfigure}{0.33\textwidth}
                    \psfrag{mag}{\hspace{-3mm} \raisebox{0mm}{\footnotesize{Magnitude}}}
                	\psfrag{t}{\hspace{-3mm}\raisebox{2mm}{\footnotesize{Sample index}}}
                    \psfrag{clu}{\hspace{0mm}\raisebox{0mm}{\footnotesize{Voronoi}}}
                	\psfrag{geo}{\hspace{0mm}\raisebox{0mm}{\footnotesize{Geohash}}}
                	\psfrag{2.0}{\hspace{0mm}\raisebox{0mm}{\footnotesize{1}}}
                	\psfrag{3.5}{\hspace{0mm}\raisebox{0mm}{\footnotesize{}}}
                	\psfrag{5.0}{\hspace{0mm}\raisebox{0mm}{\footnotesize{1440}}}
                	\psfrag{-2}{\hspace{0mm}\raisebox{0mm}{\footnotesize{-2}}}
                	\psfrag{-1}{\hspace{0mm}\raisebox{0mm}{\footnotesize{-1}}}
                	\psfrag{0}{\hspace{0mm}\raisebox{0mm}{\footnotesize{0}}}
                	\psfrag{1}{\hspace{0mm}\raisebox{0mm}{\footnotesize{1}}}
                    \includegraphics[height = 2in]{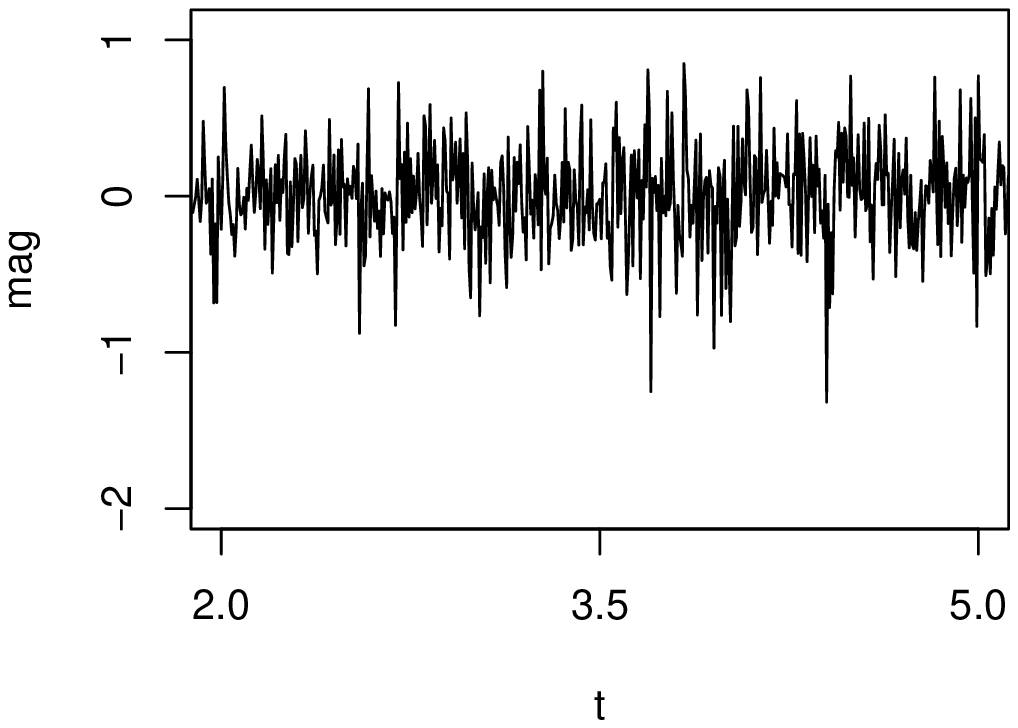}
                    \vspace{-0.5cm}
                    \caption{Resemblance to white noise}
                \end{subfigure}%
                \vspace{3mm}
    \caption{Plots obtained by performing residual diagnostics for model evaluation.}
    \label{res_diag}
 \end{figure*}

The inferences and comparisons made above are based on the spatial aggregation of the data. In the next section, we perform comparisons based on the temporal aggregated data. The spatially aggregated data for each Voronoi and Geohash partition is temporally aggregated and modeled using time-series techniques.

\section{Time-series modeling}\label{modelling}
Before modeling the data, we remove duplicate user IDs, if they appear multiple times in a 30 minute interval per partition. We assume that it is highly unlikely for a passenger to book multiple taxis within that time, from the same cell. A Box-Cox transformation is applied to stabilize the variance of the raw data, thus making the data amenable for linear processing \cite{hyndman2014forecasting}. Each K-Means demand centroid has two time-series associated with it; one using the demand aggregated at the Voronoi cell level, and the other with the demand aggregated at the geohash level. When these sequences are plotted, we observe that the series shows significant trend and seasonal patterns. On performing a spectral analysis, we find strong daily and weekly seasonality. A computationally efficient approach to real time forecasting is to use linear time series models. Hence, the data is trained using single and double-seasonal linear parametric time-series models. The city comprises of various activity zones such as residential, entertainment, office, school zones, etc. The demand patterns for each of these zones differ and hence, a single time-series model may not be a satisfactory fit for all the time-sequences. We perform the time-series modelling exercise for each tessellation strategy separately, at two time scales: 15 minutes and 60 minutes. 
\subsection{Shortlisted models} \label{models}
Holt-Winters, Auto-Regressive and Seasonal Auto-regressive Integrated Moving Average (ARIMA and SARIMA) models, Seasonal and Trend Decomposition using LOESS (STL) combined with non-seasonal exponential smoothing~\cite[Chapters 6-8] {hyndman2014forecasting} are some of the widely used single-seasonal exponential smoothing models. Double Seasonal Holt-Winters (DSHW) ~\cite{hyndman2014forecasting} and Trigonometric BATS (TBATS)~\cite{de2011forecasting} are some common double-seasonal models. In order to make sure that these models work better than n$\ddot{\text{a}}$ive alternatives, we compare the aforementioned models against a simple averaging model, n$\ddot{\text{a}}$ive and seasonal n$\ddot{\text{a}}$ive model, and random drift model. It was observed that in a few cells, the simple alternatives indeed performed better than the parametric time-series models. We consider a simple seasonal averaging model as the baseline model for our analysis. If a model performs better than the baseline model, it is kept; else it is discarded. Below, we briefly explain the models with which almost all the activity zones are well-modeled for our data.
\begin{enumerate}
    \item \textbf{Baseline model}: In order to forecast for a particular time step, the mean of all the previous season samples corresponding to that time step is computed. If weekly forecast is considered, forecast for Sunday 12 noon is the average of all Sunday 12 noon demand in the past.
    \begin{equation}
    y_{t+1} = \frac{1}{N} \sum\limits_{i = 1}^N y_{t+1 - im},
    \end{equation}
    where $y_t$ is the demand at time $t$, $N$ is the number of seasons and $m$ is the seasonality period.
        \item \textbf{TBATS}: TBATS model is a state space model introduced in ~\cite{de2011forecasting} for forecasting time-series with multiple seasonal periods, high frequency seasonal periods, non integer seasonality, and calender effects. TBATS is an acronym for Trigonometric (T), Box-Cox transform (B), ARMA errors $d_t$ (A), Trend $b_t$ (T) and Seasonal components $s^{(i)}_{t}$ (S). The seasonal components are represented using trigonometric fourier series. The equations for a $h$-step ahead additive trend, multiplicative seasonality TBATS model prediction are as follows:  
    \begin{equation}
    \begin{aligned}
    & \text{Forecast} \quad \hat{y}_{t+h|t}=(l_{t}+hb_{t})\prod_{i} s_{t-m_i+h}^{(i)}, \\
    & \text{Level} \quad l_{t}=\alpha \Bigg(\frac{y_{t}}{\prod_{i} s_{t-m_i+h}^{(i)}}\Bigg)+(1-\alpha)(l_{t-1}+b_{t-1}),\\
    & \text{Trend} \quad b_{t}=\beta(l_{t}-l_{t-1}) + (1-\beta)(b_{t-1}),\\
    &  \text{Seasonal component}\ s_{t}^{(i)}= \sum_{i=1}^{T}s_{j,t}^{i},\\
    & \text{where,}\ s_{j,t}^{(i)}=  s_{j,t-1}^{i}{\cos\lambda_{j}}^{i}+s_{j,t-1}^{*i}{\sin\lambda_{j}}^{i}+\gamma^{i}, \\
    & \ \ \ \ \ \ \ \ \lambda_{j}^{i}=\frac{2\pi j}{m_{i}}. 
    \end{aligned}
    \end{equation}
     where $\alpha$, $\beta$, and $\gamma$ are the smoothing parameters, and $m_{i}$, $i \in \{1, \ldots, T\}$ are the seasonal periods.
    \item \textbf{Seasonal and Trend decomposition using Loess (STL)}: STL is a time-series decomposition technique where the data is decomposed into seasonal and non-seasonal components. The non seasonal component is modeled using techniques such as ARIMA, exponential smoothing models (ETS), random drift etc. The seasonal component, after a Seasonal N$\ddot{\text{a}}$ive forecast for the seasons, is added to the non-seasonal component to form the final model.
    \begin{subequations}
    \begin{equation}
    y_t = {S_t} + {NS_t},
    \end{equation}
    \begin{equation}
    {NS_t} = {T_t}+ {R_t},
    \end{equation}
    \begin{equation}
    {S_{T+h}} = S_{T+h-km};\quad k = \lfloor (h-1)/m\rfloor+1.
    \end{equation}
    \end{subequations}
    Here, $T_t$ denotes the trend, $R_t$ is the irregular component, $\lfloor u \rfloor$ is the integer part of variable $u$, $h$ is the prediction horizon, and $m$ is the seasonal period.
\end{enumerate}

\subsection{Model validation}
 The suitability of the models for the data can be analyzed as follows:
\begin{itemize}
    \item Plotting the histogram of the residuals: The histogram should follow a Gaussian distribution with residual mean as the mean of the distribution.
    \item Plotting the Auto Correlation Function (ACF) of the residuals: The residuals should lie within the significant band of the ACF plot, i.e, within $\pm 2/\sqrt{N}$ band, where $N$ is the sample size.
    \item Resemblance with white noise: The residuals should appear to be uncorrelated and should pass a statistical test for correlation; for eg., the Box-Ljung test. The Box-Ljung statistic is a function of the accumulated sample auto correlations up to any specified time lag.
\end{itemize}
We refer the reader to ~\cite[Chapter 2, Section $2.6$] {hyndman2014forecasting} for an elaborate explanation of the afore-mentioned residual diagnostic tools. Results of residual diagnostics performed on the Voronoi time-series generated from an entertainment zone in Bengaluru are plotted in Figure \ref{res_diag}. Here, STL decomposition with ARIMA(2,1,2) performed satisfactorily and the residuals from the fit appear to be uncorrelated. 
\begin{table*}  
\parbox{0.5\textwidth}{
\scalebox{0.95}{
\centering
\begin{tabular}{|c||c||c|}
\hline
\textbf{Strategy} & \textbf{Bengaluru}                                                                                      & \textbf{New York}                                                                                    \\ \hline \hline
Geohash  & \begin{tabular}[c]{@{}c@{}}STL(ETS,a.e,d.t),\\ STL(ARIMA, BoxCox)\end{tabular}                 & \begin{tabular}[c]{@{}c@{}}STL(ETS,a.e,d.t),\\ STL(randomdrift)\end{tabular}                \\ \hline
Voronoi  & \begin{tabular}[c]{@{}c@{}}STL(ETS,a.e,d.t),\\ TBATS(ARMA error, \\ BoxCox,trend)\end{tabular} & \begin{tabular}[c]{@{}c@{}}STL(ETS,a.e,d.t),\\ STL(ARIMA,BoxCox),\\ STL(naive)\end{tabular} \\ \hline
\end{tabular}}
\caption{Best performing models for the datasets, \\ where a.e = additive errors, and d.t = damped trend.}
\label{best_models}
}\hfill
\parbox{0.5\textwidth}{
\scalebox{0.95}{
\centering
\begin{tabular}{|c||l||c|c|c|c|}
\hline
\multirow{2}{*}{\textbf{Data set}} & \multicolumn{1}{c|}{\multirow{2}{*}{\textbf{\begin{tabular}[c]{@{}c@{}}Error\\ Metric\end{tabular}}}} & \multicolumn{2}{c|}{60 min.}                                   & \multicolumn{2}{c|}{15 min.}                                            \\ \cline{3-6} 
                                   & \multicolumn{1}{c|}{}                                                                                 & Vor                       & Geo                                & Vor                                & Geo                                \\ \hline \hline
\multirow{2}{*}{Bengaluru}         & SMAPE (\%)                                                                                            & \textbf{17.6}             & 21.7                               & \textbf{37}                        & 40.4                               \\ \cline{2-6} 
                                   & MASE                                                                                                  & \multicolumn{1}{l|}{0.75} & \multicolumn{1}{l|}{\textbf{0.73}} & \multicolumn{1}{l|}{\textbf{0.52}} & \multicolumn{1}{l|}{0.53}          \\ \hline
\multirow{2}{*}{New York}          & SMAPE (\%)                                                                                            & \textbf{15.2}             & 15.6                               & 33.6                               & \textbf{31.2}                      \\ \cline{2-6} 
                                   & MASE                                                                                                  & \multicolumn{1}{l|}{0.60} & \multicolumn{1}{l|}{\textbf{0.59}} & \multicolumn{1}{l|}{0.62}          & \multicolumn{1}{l|}{\textbf{0.47}} \\ \hline
\end{tabular}}
\caption{Preliminary results for the Bengaluru and New York data sets.}
\label{table_prelim}
}
\end{table*}

The performance of the model can further be evaluated by employing performance metrics such as SMAPE and MASE. The SMAPE is a symmetrized version of the Mean Absolute Percent Error, which is defined if, at all future time points, the point forecasts and actuals are both not zero. For a forecast horizon $N$ considered, SMAPE is defined as follows:
\begin{equation}
\text{SMAPE} = \frac{100}{N} \mathlarger{\sum_{t=1}^{N}}\frac{|\hat{y}_{t}-y_{t}|}{\hat{y}_{t}+y_{t}},
\end{equation}
where $y_t$ is the actual demand and $\hat{y_t}$ is the forecast at time $t$.
Even though this metric is widely used in industry, SMAPE is a scale-dependent error, and not well-suited for intermittent demand. Hence, we also evaluate our models against Mean Absolute Scaled Error (MASE). For a time-series of forecast horizon $N$ and seasonal period $m$, MASE is defined as:
\begin{equation}
    \text{MASE} = \frac{1}{N} \mathlarger{\sum_{t= 1}^N}\Bigg (\frac{|y_t - \hat{y_t}|}{\frac{1}{N-m}\sum_{t=m+1}^N |y_t - y_{t-m}|} \Bigg ).
\end{equation}
For a non-seasonal time-series, $m = 1$. The denominator of the equation is the mean absolute error of the one-step seasonal n$\ddot{\text{a}}$ive forecast method on the training set. This error metric compares the models with the standard random drift model, and ensures that the models perform better than a n$\ddot{\text{a}}$ive technique. MASE can be used to compare forecasts across data sets with different scales, as the metric is independent of data scale. Smaller the value of the metric, the better is the model. The selected models passed the residual diagnostics and performed satisfactorily with SMAPE and MASE. In Table \ref{best_models}, we list the shortlisted models with which Bengaluru and New York are well-modelled at 60 min. aggregation levels. The errors encountered by the individual models are not listed because the modeling technique that works for one scenario might not provide good performance with another scenario.
\vspace{-1mm}
 \begin{figure}[h!]
    \begin{subfigure}[t]{0.5\linewidth}
                    \psfrag{no}{\hspace{-8mm} \raisebox{3mm}{\footnotesize{Sample size (norm.)}}}
                	\psfrag{smape}{\hspace{-1mm}\raisebox{0mm}{\scriptsize{SMAPE}}}
                  	 \psfrag{clu}{\hspace{0mm}\raisebox{0mm}{\footnotesize{Voronoi}}}
                	\psfrag{geo}{\hspace{0mm}\raisebox{0mm}{\footnotesize{Geohash}}}
                	\psfrag{0}{\hspace{-2mm} \raisebox{1mm}{\footnotesize{0}}}
                	\psfrag{275}{\hspace{-1mm} \raisebox{1mm}{\footnotesize{}}}
                	\psfrag{550}{\hspace{-2mm} \raisebox{1mm}{\footnotesize{1}}}                	
                	\psfrag{9}{\hspace{-2mm} \raisebox{-0mm}{\footnotesize{9}}}
                	\psfrag{18}{\hspace{-2mm} \raisebox{-0mm}{\footnotesize{18}}}
                	\psfrag{27}{\hspace{-2mm} \raisebox{-0mm}{\footnotesize{27}}}
    \includegraphics[height = 2in,width=\linewidth]{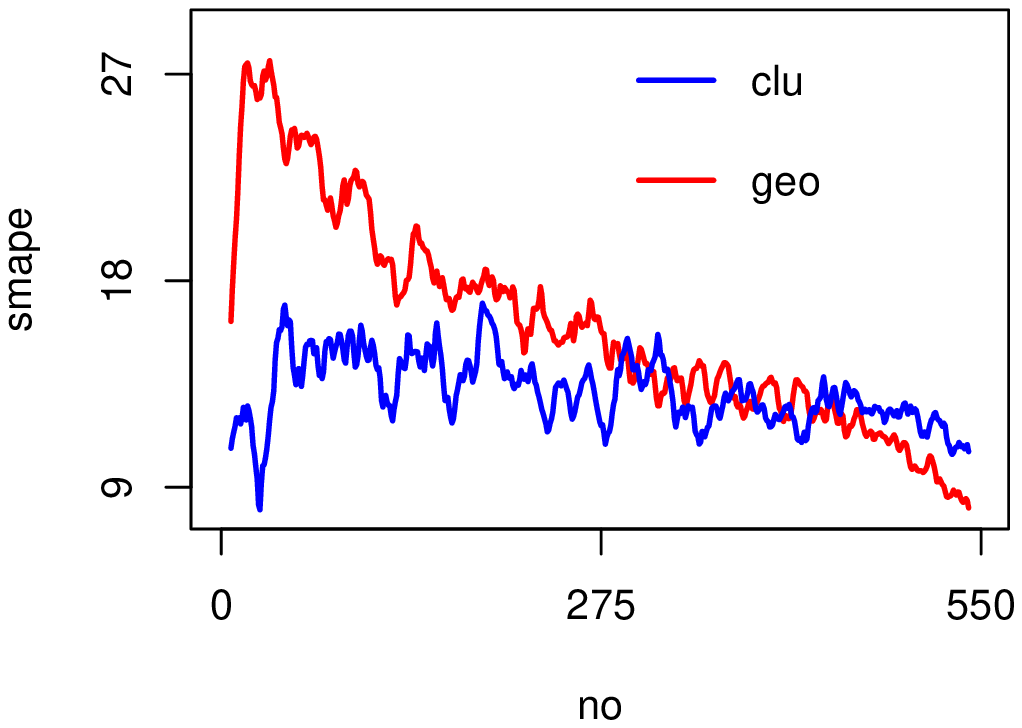}
    \vspace{-1cm}
    \caption{Bengaluru}
    \label{f1}
\end{subfigure}%
    \hfill%
\begin{subfigure}[t]{0.5\linewidth}
                            \psfrag{nopoints}{\hspace{-8mm} \raisebox{3mm}{\footnotesize{Sample size (norm.)}}}
                	\psfrag{smape}{\hspace{-1mm}\raisebox{0mm}{\scriptsize{SMAPE}}}
                  	 \psfrag{clu}{\hspace{0mm}\raisebox{0mm}{\footnotesize{Voronoi}}}
                	\psfrag{geo}{\hspace{0mm}\raisebox{0mm}{\footnotesize{Geohash}}}
                	    	\psfrag{4}{\hspace{-2mm} \raisebox{-0mm}{\footnotesize{4}}}
                	\psfrag{11}{\hspace{-2mm} \raisebox{-0mm}{\footnotesize{11}}}
                	\psfrag{18}{\hspace{-2mm} \raisebox{-0mm}{\footnotesize{18}}}
                	  	\psfrag{0}{\hspace{-2mm} \raisebox{1mm}{\footnotesize{0}}}
                	\psfrag{1}{\hspace{-2mm} \raisebox{1mm}{\footnotesize{1}}} 
    \includegraphics[height = 2in,width=\linewidth]{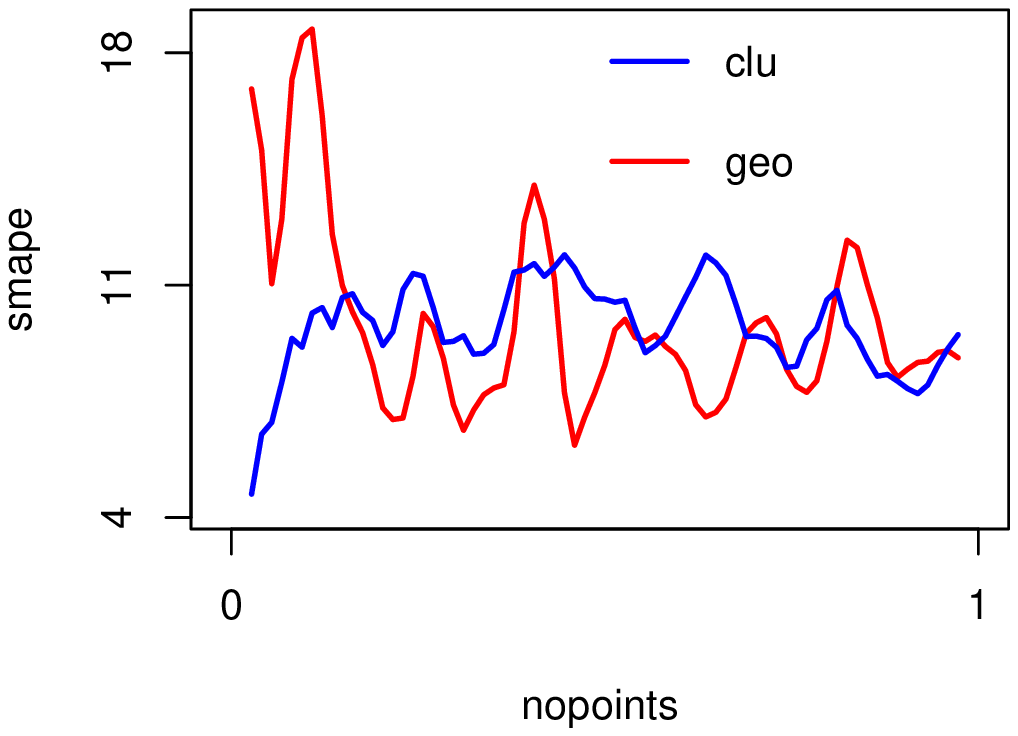}
    \vspace{-1cm}
    \caption{Manhattan-Bronx}
    \label{f2}
\end{subfigure}
\caption{The variation of SMAPE with the partition size at 60 minute aggregation level.}
  \label{nopoints_smape}
\end{figure}
\subsection{Preliminary results}\label{results}
After tessellating the city space into geohashes and Voronoi cells, we ran the time-series models mentioned in the previous section on the spatially and temporally aggregated area-normalized demand. In general, We observe strong heterogeneity in city demand across both spatial and temporal dimensions. It was observed that for cells with low density, Voronoi tessellation performs better than Geohash tessellation (Figure \ref{nopoints_smape}). The optimal density based partition in Voronoi cells appears to be the reason for this behavior. For high density cells, Geohash based models perform at least as good as Voronoi based models. The comparable performance along with the low computational complexity makes Geohash tessellation the preferred choice for data dense locations. The numerical results are summarized below:
\begin{enumerate}
    \item Using the models mentioned in Section \ref{models}, we achieve an average accuracy of about 80\% per $\text{km}^2$ for both data sets, at 60 min. aggregation levels.
    \item On referring Table \ref{table_prelim} for the performance evaluation of Geohash and Voronoi tessellation based models on the 740 Bengaluru city and 780 New York city demand centers, we notice the absence of an universal winner. 
    \item On average, Voronoi tessellation appears to outperform Geohash tessellation for Bengaluru city. On the other hand, Geohash tessellation emerges as the superior technique for most of the tested use cases for the New York city.   
\end{enumerate}
 \begin{figure*}
                \begin{subfigure}{0.25\textwidth}
                    \psfrag{smape}{\hspace{-3mm} \raisebox{3mm}{\scriptsize{SMAPE}}}
                	\psfrag{cdf}{\hspace{-3mm}\raisebox{0mm}{\footnotesize{CDF}}}
                    \psfrag{c1}{\hspace{0mm}\raisebox{0mm}{\footnotesize{V-60}}}
                	\psfrag{g1}{\hspace{0mm}\raisebox{0mm}{\footnotesize{G-60}}}                    \psfrag{c2}{\hspace{0mm}\raisebox{0mm}{\footnotesize{V-15}}}
                	\psfrag{g2}{\hspace{0mm}\raisebox{0mm}{\footnotesize{G-15}}}
                	\psfrag{0.0}{\hspace{0mm}\raisebox{0mm}{\footnotesize{0}}}
                	\psfrag{0.5}{\hspace{0mm}\raisebox{0mm}{\footnotesize{0.5}}}
                	\psfrag{1.0}{\hspace{0mm}\raisebox{0mm}{\footnotesize{1}}}
                	\psfrag{0}{\hspace{0mm}\raisebox{1mm}{\footnotesize{0}}}
                	\psfrag{20}{\hspace{0mm}\raisebox{1mm}{\footnotesize{20}}}
                	\psfrag{40}{\hspace{0mm}\raisebox{1mm}{\footnotesize{40}}}
                    \includegraphics[height = 2in, width=\linewidth]{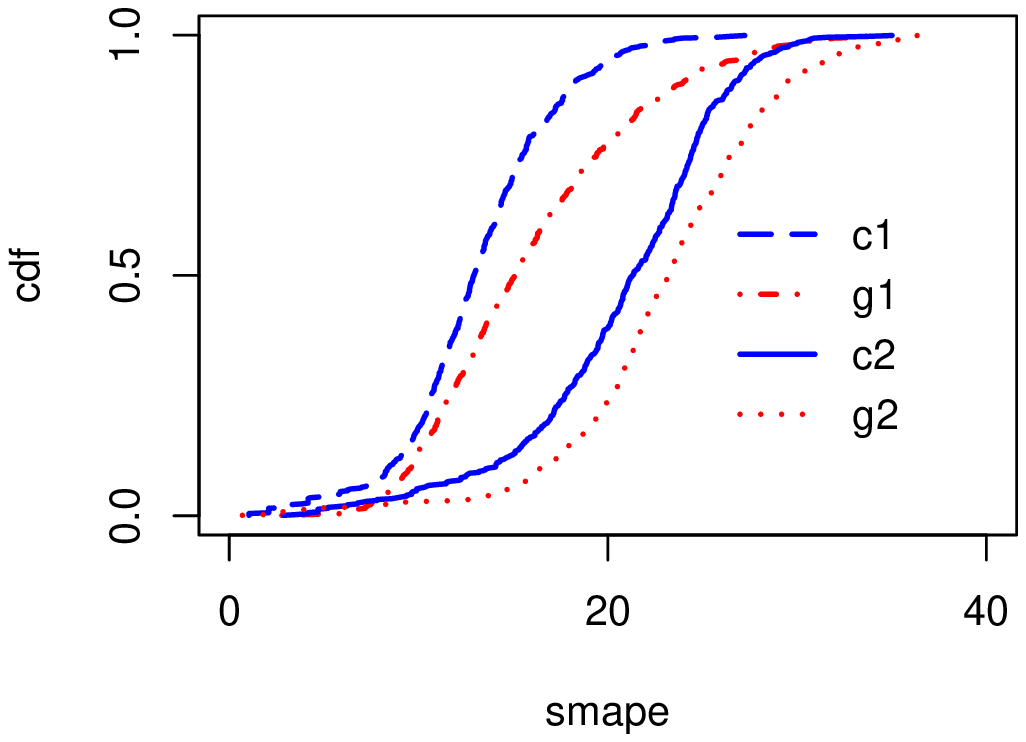}
                    \vspace{-1.1cm}
                    \caption{Bengaluru}
                    \label{smape_bang_ecdf}
                \end{subfigure}%
                \begin{subfigure}{0.25\textwidth}
                    \psfrag{smape}{\hspace{-3mm} \raisebox{3mm}{\scriptsize{MASE}}}
                	\psfrag{cdf}{\hspace{-3mm}\raisebox{0mm}{\footnotesize{CDF}}}
                    \psfrag{c1}{\hspace{0mm}\raisebox{0mm}{\footnotesize{V-60}}}
                	\psfrag{g1}{\hspace{0mm}\raisebox{0mm}{\footnotesize{G-60}}}                    \psfrag{c2}{\hspace{0mm}\raisebox{0mm}{\footnotesize{V-15}}}
                	\psfrag{g2}{\hspace{0mm}\raisebox{0mm}{\footnotesize{G-15}}}
                	\psfrag{0.0}{\hspace{0mm}\raisebox{0mm}{\footnotesize{0}}}
                	\psfrag{0.5}{\hspace{0mm}\raisebox{0mm}{\footnotesize{0.5}}}
                	\psfrag{1.0}{\hspace{0mm}\raisebox{0mm}{\footnotesize{1}}}                	
                	\psfrag{0l}{\hspace{0mm}\raisebox{1mm}{\footnotesize{0}}}
                	\psfrag{1}{\hspace{0mm}\raisebox{1mm}{\footnotesize{1}}}
                	\psfrag{2}{\hspace{0mm}\raisebox{1mm}{\footnotesize{2}}}
                    \includegraphics[height = 2in, width=\linewidth]{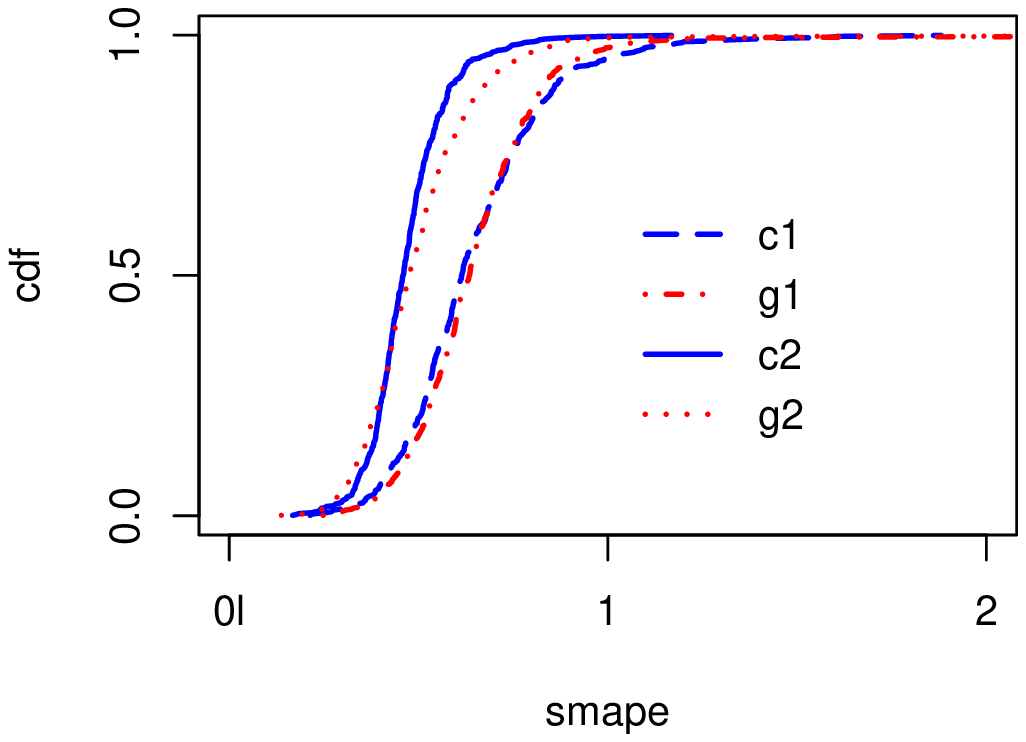}
                    \vspace{-1.1cm}
                    \caption{Bengaluru}
                    \label{mase_bang_ecdf}
                \end{subfigure}%
                    \begin{subfigure}{0.25\textwidth}
                    \psfrag{smape}{\hspace{-3mm} \raisebox{3mm}{\scriptsize{SMAPE}}}
                	\psfrag{cdf}{\hspace{-3mm}\raisebox{0mm}{\footnotesize{CDF}}}
                    \psfrag{c1}{\hspace{0mm}\raisebox{0mm}{\footnotesize{V-60}}}
                	\psfrag{g1}{\hspace{0mm}\raisebox{0mm}{\footnotesize{G-60}}}                    \psfrag{c2}{\hspace{0mm}\raisebox{0mm}{\footnotesize{V-15}}}
                	\psfrag{g2}{\hspace{0mm}\raisebox{0mm}{\footnotesize{G-15}}}
                	\psfrag{0.0}{\hspace{0mm}\raisebox{0mm}{\footnotesize{0}}}
                	\psfrag{0.5}{\hspace{0mm}\raisebox{0mm}{\footnotesize{0.5}}}
                	\psfrag{1.0}{\hspace{0mm}\raisebox{0mm}{\footnotesize{1}}}
                	\psfrag{0l}{\hspace{0mm}\raisebox{1mm}{\footnotesize{0}}}
                	\psfrag{20}{\hspace{-1mm}\raisebox{1mm}{\footnotesize{20}}}
                	\psfrag{40}{\hspace{-1mm}\raisebox{1mm}{\footnotesize{40}}}
                    \includegraphics[height = 2in, width=\linewidth]{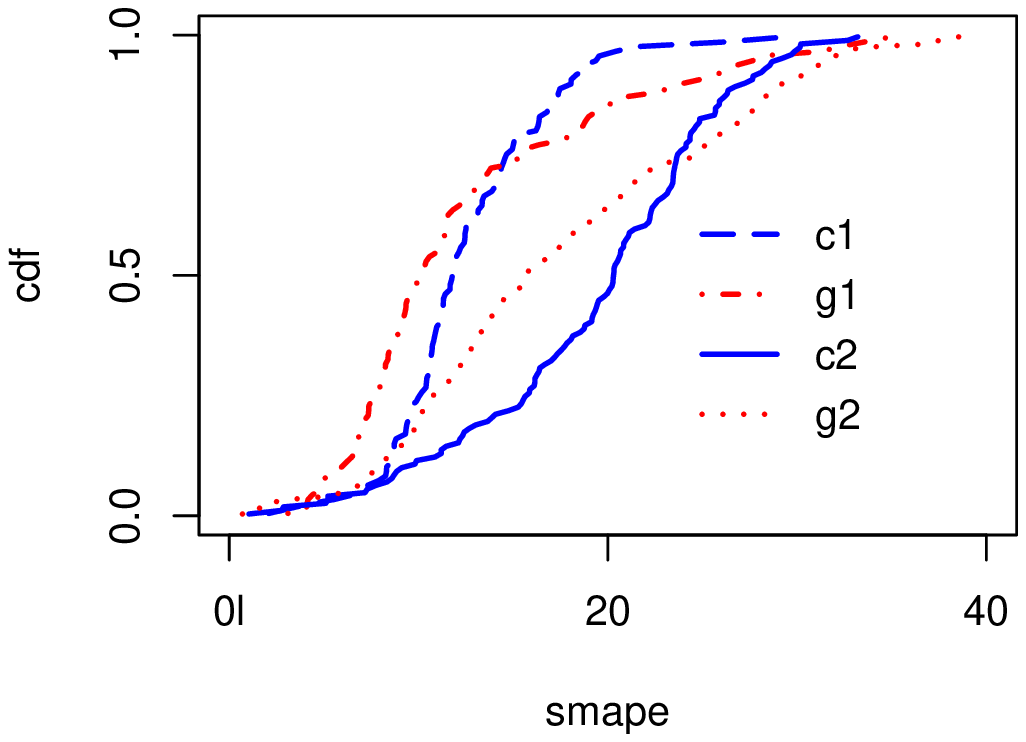}
                    \vspace{-1.1cm}
                    \caption{New York}
                    \label{smape_nyc_ecdf}
                \end{subfigure}%
                \begin{subfigure}{0.25\textwidth}
                    \psfrag{smape}{\hspace{-3mm} \raisebox{3mm}{\scriptsize{MASE}}}
                	\psfrag{cdf}{\hspace{-3mm}\raisebox{0mm}{\footnotesize{CDF}}}
                    \psfrag{c1}{\hspace{0mm}\raisebox{0mm}{\footnotesize{V-60}}}
                	\psfrag{g1}{\hspace{0mm}\raisebox{0mm}{\footnotesize{G-60}}}                    \psfrag{c2}{\hspace{0mm}\raisebox{0mm}{\footnotesize{V-15}}}
                	\psfrag{g2}{\hspace{0mm}\raisebox{0mm}{\footnotesize{G-15}}}
                   	\psfrag{0.0}{\hspace{0mm}\raisebox{0mm}{\footnotesize{0}}}
                	\psfrag{0.5}{\hspace{0mm}\raisebox{0mm}{\footnotesize{0.5}}}
                	\psfrag{1.0}{\hspace{0mm}\raisebox{0mm}{\footnotesize{1}}}                	
                	\psfrag{0l}{\hspace{0mm}\raisebox{1mm}{\footnotesize{0}}}
                	\psfrag{0.75}{\hspace{0mm}\raisebox{1mm}{\footnotesize{0.75}}}
                	\psfrag{1.5}{\hspace{-1mm}\raisebox{1mm}{\footnotesize{1.5}}}
                    \includegraphics[height = 2in, width=\linewidth]{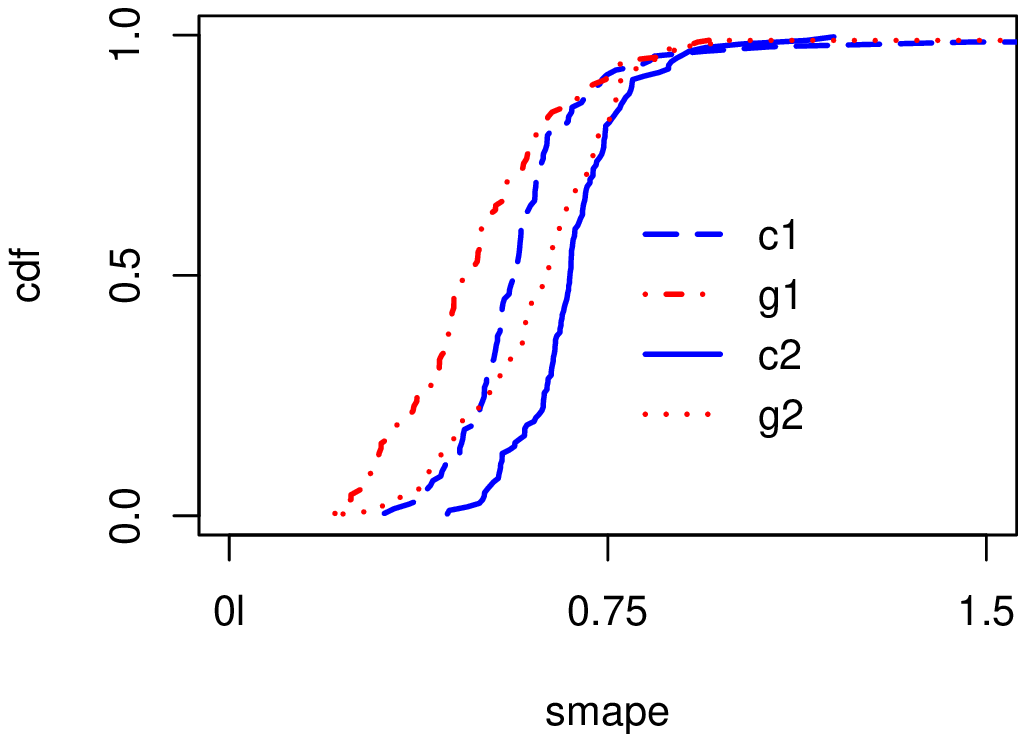}
                    \vspace{-1.1cm}
                    \caption{New York}
                    \label{mase_nyc_ecdf}
                \end{subfigure}%
                \vspace{3mm}
    \caption{ECDF plots of Voronoi and Geohash partition techniques using MASE and SMAPE as the metrics. V-XX and G-XX correspond to performance of Voronoi and Geohash based models at XX min. aggregation levels.}
    \label{ecdfs}
 \end{figure*} 
 \begin{figure*}
                \begin{subfigure}{0.33\textwidth}
                	\psfrag{t}{\hspace{-3mm}\raisebox{3mm}{\footnotesize{Time step}}}
                	\psfrag{mase}{\hspace{-3mm}\raisebox{-2mm}{\footnotesize{SMAPE}}}
                	 \psfrag{clu}{\hspace{0mm}\raisebox{0mm}{\footnotesize{Voronoi}}}
                	\psfrag{geo}{\hspace{0mm}\raisebox{0mm}{\footnotesize{Geohash}}}                	
                	\psfrag{1}{\hspace{-1mm} \raisebox{1mm}{\footnotesize{1}}}                	
                	\psfrag{22}{\hspace{-1mm} \raisebox{-1mm}{\footnotesize{22}}}
                	\psfrag{48}{\hspace{-1mm} \raisebox{1mm}{\footnotesize{48}}}
                	\psfrag{53}{\hspace{-1mm} \raisebox{-1mm}{\footnotesize{53}}}                	
                	\psfrag{96}{\hspace{-1mm} \raisebox{1mm}{\footnotesize{96}}}                	
                	\psfrag{84}{\hspace{-1mm} \raisebox{-1mm}{\footnotesize{84}}}
                    \includegraphics[height = 2in]{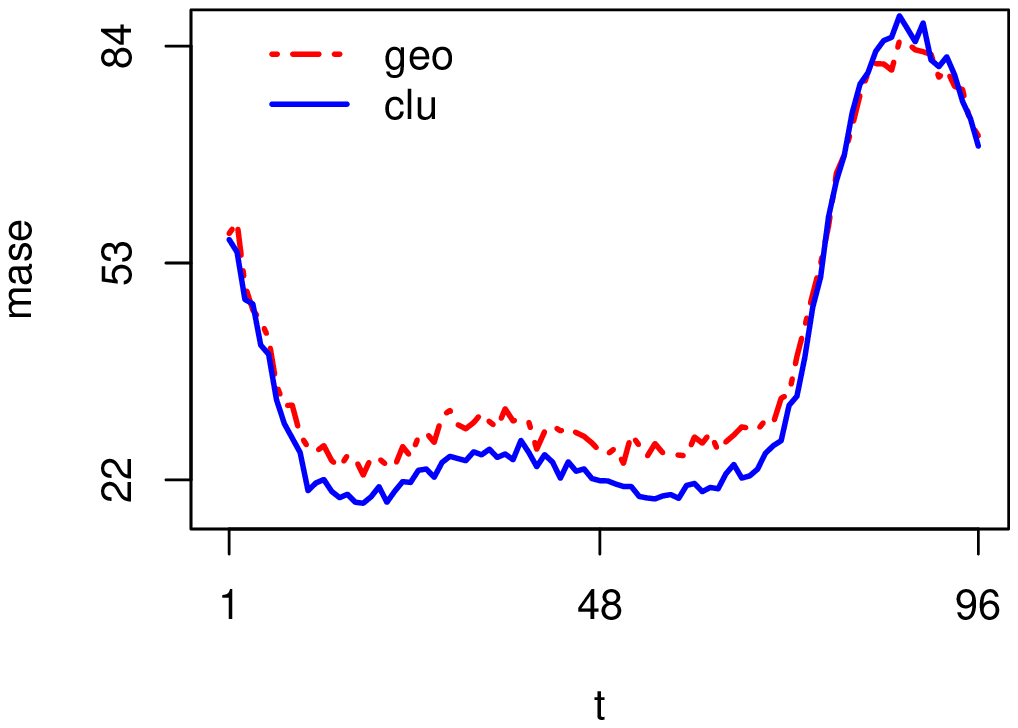}
                        \vspace{-0.65cm}
                    \caption{Without HEDGE}
                    \label{before_hedge}
                \end{subfigure}%
                \begin{subfigure}{0.33\textwidth}
                    \psfrag{t}{\hspace{-7mm} \raisebox{3mm}{\footnotesize{Time step}}}
                	\psfrag{mase}{\hspace{-3mm}\raisebox{-2mm}{\footnotesize{SMAPE}}}             \psfrag{clu}{\hspace{0mm}\raisebox{0mm}{\footnotesize{Voronoi}}}
                	\psfrag{geo}{\hspace{0mm}\raisebox{0mm}{\footnotesize{Geohash}}} 
                	\psfrag{hedge}{\hspace{0mm}\raisebox{0mm}{\footnotesize{HEDGE}}} 
                  \psfrag{1}{\hspace{-1mm} \raisebox{1mm}{\footnotesize{1}}}                	
                	\psfrag{22}{\hspace{-1mm} \raisebox{-1mm}{\footnotesize{22}}}
                	\psfrag{48}{\hspace{-1mm} \raisebox{1mm}{\footnotesize{48}}}
                	\psfrag{53}{\hspace{-1mm} \raisebox{-1mm}{\footnotesize{53}}}                	
                	\psfrag{96}{\hspace{-1mm} \raisebox{1mm}{\footnotesize{96}}}                	
                	\psfrag{84}{\hspace{-1mm} \raisebox{-1mm}{\footnotesize{84}}}
                   \includegraphics[height = 2in]{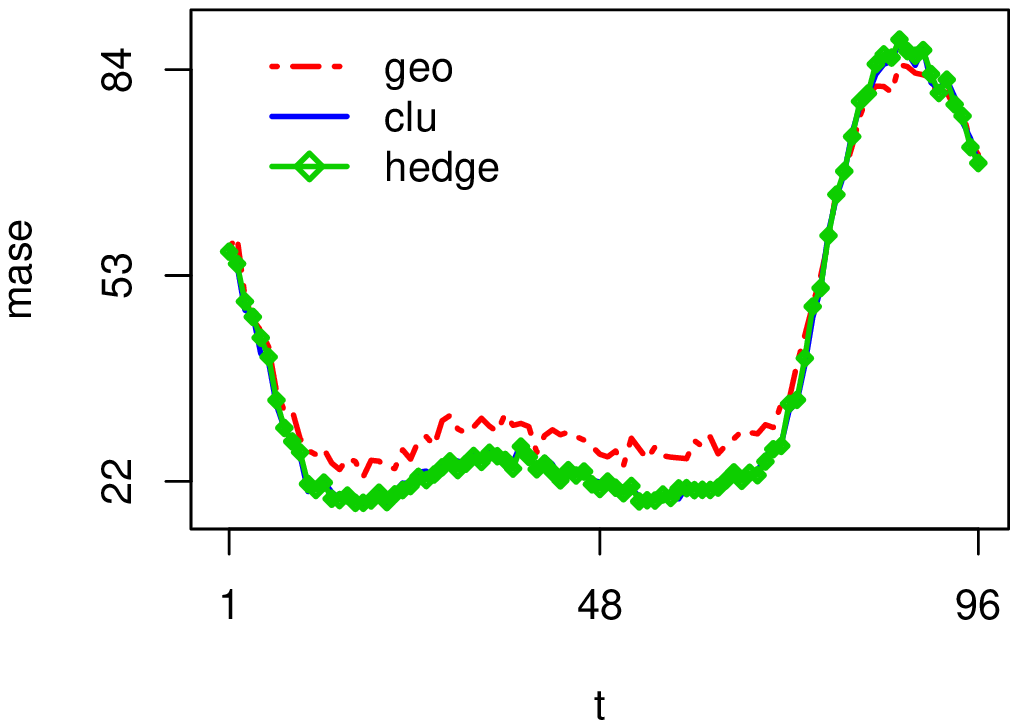}
                        \vspace{-0.65cm}
                    \caption{HEDGE without discounting}
                    \label{after_hedge_nodisc}
                \end{subfigure}%
                \begin{subfigure}{0.33\textwidth}
                    \psfrag{t}{\hspace{-7mm} \raisebox{3mm}{\footnotesize{Time step}}}
                	\psfrag{mase}{\hspace{-3mm}\raisebox{-2mm}{\footnotesize{SMAPE}}}
                	 \psfrag{clu}{\hspace{0mm}\raisebox{0mm}{\footnotesize{Voronoi}}}
                	\psfrag{geo}{\hspace{0mm}\raisebox{0mm}{\footnotesize{Geohash}}}
                	 \psfrag{h2}{\hspace{0mm}\raisebox{0mm}{\footnotesize{dHEDGE}{(0.1,0.7)}}} 
                \psfrag{1}{\hspace{-1mm} \raisebox{1mm}{\footnotesize{1}}}                	
                	\psfrag{22}{\hspace{-1mm} \raisebox{-1mm}{\footnotesize{22}}}
                	\psfrag{48}{\hspace{-1mm} \raisebox{1mm}{\footnotesize{48}}}
                	\psfrag{53}{\hspace{-1mm} \raisebox{-1mm}{\footnotesize{53}}}                	
                	\psfrag{96}{\hspace{-1mm} \raisebox{1mm}{\footnotesize{96}}}                	
                	\psfrag{84}{\hspace{-1mm} \raisebox{-1mm}{\footnotesize{84}}}
                    \includegraphics[height = 2in]{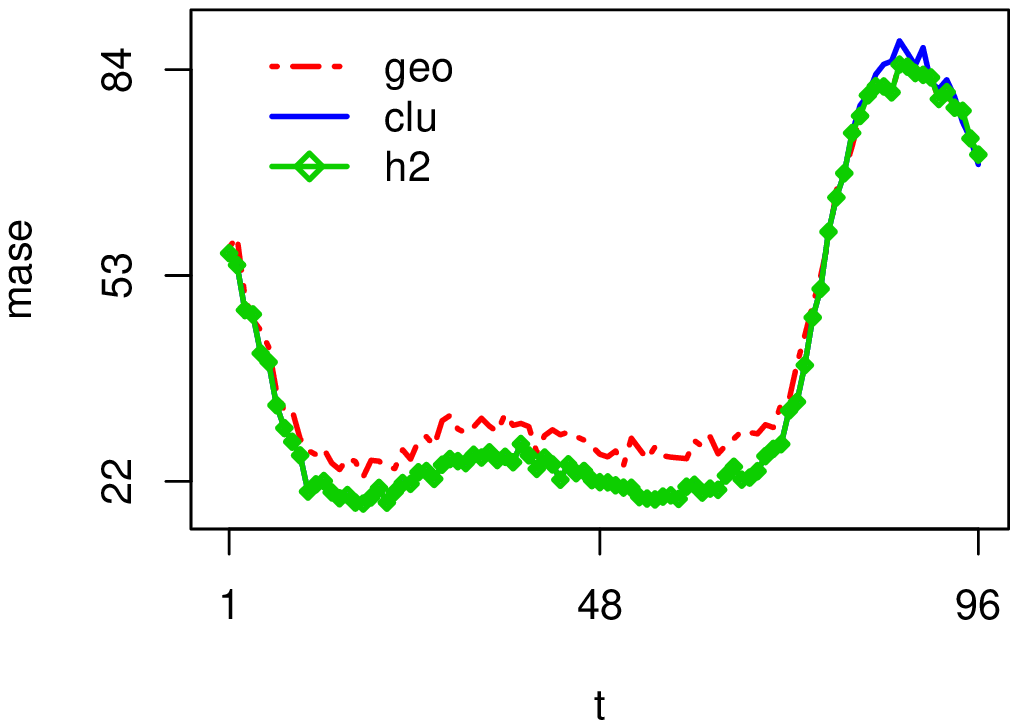}
                       \vspace{-0.65cm}
                    \caption{HEDGE with discounting}
                    \label{after_hedge_disc}
                \end{subfigure}%
                \vspace{2mm}
    \caption{The performances of models based on tessellation techniques without HEDGE, after original HEDGE, and after HEDGE with discounting.}
 \end{figure*}

Further, the Empirical Cumulative Distribution Function (ECDF) of the average errors obtained from the models based on all the Voronoi and Geohash partitions are plotted in Figure \ref{ecdfs}. We can see a dominance of Voronoi technique over Geohash technique at 15 and 60 minute aggregation levels in Figures \ref{smape_bang_ecdf} and \ref{mase_bang_ecdf} for Bengaluru city. The errors from Voronoi based models reach the upper bound faster than errors from Geohash based models. On the other hand, for New York city, Figures \ref{smape_nyc_ecdf} and \ref{mase_nyc_ecdf} show a dominance of Geohash technique over Voronoi technique. In addition, we conduct the Kolmogorov-Smirnov test (K-S test) \cite{massey1951kolmogorov} to check if the ECDFs differ significantly. The results obtained provide sufficient evidence that Voronoi technique is statistically better than Geohash technique for Bengaluru city and vice versa for New York city, consistent with the inferences from Figure \ref{ecdfs}. 
Hence, we find that Geohash tessellation performs better than Voronoi for the Bengaluru city, while Voronoi tessellation has a superior performance over Geohash for New York city. Further, we proceed to plot the instantaneous errors obtained from the models based on the two tessellation techniques in Figure \ref{before_hedge} for Bengaluru at 15 minute aggregation. Each point on the graph corresponds to the mean of errors obtained from all geohash/Voronoi cells at that time instant.  We can see that early mornings and late nights, Geohash based models work better compared to Voronoi based models. For the rest of the day, \emph{i.e.,} the daylight hours, Voronoi based models tend to produce better forecasts compared to Geohash based models. It is surprising to note that even within a single city, a unique tessellation technique does not have superior performance over the other at all time steps. This non-stationary behavior of the models prompts us to combine the models for better location-based forecasts. To that end, we propose a modified version of discounted HEDGE algorithm \cite{raj2017aggregating} in the next section.

\section{Modified HEDGE for combining models}\label{hedgealgo}
Consider a learning framework, where model provides recommendations. Each model is referred to as an $expert$. Assuming that the predictor has complete knowledge about all the past decisions and performances of experts, the goal is to perform as good as the best expert in the pool. This exercise belongs to the class of ensemble learning where we use multiple learning algorithms to obtain better predictive performance than could be obtained from any of the constituent learning algorithms alone. Here, whenever Geohash tessellation performs better than Voronoi tessellation at a time step $t$, the ideal decision is to choose Geohash technique as the tessellation strategy for that $t$. For all demand centers, the forecasts at that instant should then be made from their geohash cells.

On observing the performances of both strategies at 15 and 60 minute aggregation levels, we see that a single tessellation technique is unable to yield optimal results for the entire forecast horizon. The best strategy varies with time. A sensible approach towards solving this issue is to combine the two experts. As an initial step towards that goal, we apply the classic HEDGE algorithm as proposed in \cite{freund1995desicion} to combine the models (Figure \ref{after_hedge_nodisc}). We see that the decision maker is able to follow the best expert, model based on Voronoi in this case, till a cross over occurs. The algorithm is unable to adapt to the shifting nature of the experts. In order to adapt to such a non stationary environment, dHEDGE \cite{raj2017aggregating} was proposed, where the authors modify the conventional HEDGE to include an exponential discounting factor. By exponentially filtering the past performances of the experts, dHEDGE takes into account the non-stationarity of the process. By normalizing the weights of each expert at every instant, dHEDGE provides a set of coefficients for experts. This can then be used to generate a convex combination of expert opinion.  Our problem do not require normalized weights. Since there are only two experts, at every instant we can pick the expert which has maximum weight, $i.e$, the expert with the minimum loss. Hence, we modify the dHEDGE algorithm to pick an expert if it has a higher weight than the other expert. The original algorithm has a generic loss function $L(x,y) \in [0,1]$. For illustrative purposes, they have used a discrete loss function $L(x,y) = [[x \neq y]]$, where $[[\cdot]]$ is the indicator function, $x$ is the observed symbol and $y$ is the predicted symbol. For their cellular LTE network application, that loss function outputs a $0$ for all the experts who predicted the symbol correctly at that instant, and output a $1$ for all other experts. On the other hand, we build our loss functions based on the errors observed for each strategy. Our modified dHEDGE algorithm is given in Algorithm \ref{hedge}.
\begin{algorithm}
\setstretch{1.35}
    \textbf{Parameters:} Learning para. $\beta \in [0,1]$, Discounting factor $\gamma \in [0,1]$ \\
    \textbf{Initialization:} Set $w_k[1] = W> 0$ for $k \in {1,2}$ s.t $\sum_{i = 1}^2 w_i[1] = 1$ \\
      \For{$t = 1,2, \ldots$}
       {Choose expert with index = $\underset{i}{argmax}(w_i[t])$ \\
        Error $e_v$ = MEAN(forecast errors from models based on Voronoi at $t$) \\
        Error $e_g$ = MEAN(forecast errors from models based on Geohash at $t$)\\
        Loss L at $t$, $l_1[t] = \frac{e_v}{e_v + e_g}$, $l_2[t] = \frac{e_g}{e_v + e_g}$ \\
        Update weights as $w_i[t+1] = w_i[t]^{\gamma} \cdot \beta^{l_i[t]}$ 
      }
   \caption{Modified dHEDGE for choosing the best expert}\label{hedge}
\end{algorithm}
\begin{figure*}[htbp!]
                \begin{subfigure}{\textwidth}
                    \psfrag{t}{\hspace{-7mm} \raisebox{2mm}{\footnotesize{Time step}}}
                	\psfrag{mase}{\hspace{-3mm}\raisebox{-3mm}{\scriptsize{SMAPE}}}
                	\psfrag{cms}{\hspace{-8mm}\raisebox{-1mm}{\footnotesize{Cum. Mean Error}}}
                     \psfrag{clu}{\hspace{0mm}\raisebox{0mm}{\scriptsize{Voronoi}}}
                	\psfrag{geo}{\hspace{0mm}\raisebox{0mm}{\scriptsize{Geohash}}}                	 
                	\psfrag{h1}{\hspace{0mm}\raisebox{-2mm}{\shortstack{\scalebox{0.65}{dHEDGE}\\\scalebox{0.5}{(0.1,0.1)}}}}     
                	\psfrag{h2}{\hspace{-1mm}\raisebox{-2mm}{\shortstack{\scalebox{0.65}{dHEDGE}\\\scalebox{0.5}{(0.1,0.4)}}}}  
                	\psfrag{h3}{\hspace{0mm}\raisebox{-2mm}{\shortstack{\scalebox{0.65}{dHEDGE}\\\scalebox{0.5}{(0.1,0.1)}}}}        
                	\psfrag{h4}{\hspace{0mm}\raisebox{-2mm}{\shortstack{\scalebox{0.65}{dHEDGE}\\\scalebox{0.5}{(0.1,0.4)}}}}     
                	\psfrag{11}{\hspace{-1mm} \raisebox{-0.5mm}{\footnotesize{11}}}                	
                	\psfrag{19}{\hspace{-1mm} \raisebox{-0.5mm}{\footnotesize{19}}}
                	\psfrag{51}{\hspace{-2mm} \raisebox{-0.5mm}{\footnotesize{51}}} 	
                	\psfrag{88}{\hspace{-2mm} \raisebox{-0.5mm}{\footnotesize{88}}}                	
                	\psfrag{13}{\hspace{-1mm} \raisebox{-0.5mm}{\footnotesize{13}}}
                	\psfrag{27}{\hspace{-1mm} \raisebox{-0.5mm}{\footnotesize{27}}}
                	\psfrag{25}{\hspace{-1mm} \raisebox{-0.5mm}{\footnotesize{25}}}                	
                	\psfrag{58}{\hspace{-2mm} \raisebox{-0.5mm}{\footnotesize{58}}}
                	\psfrag{42}{\hspace{-1mm} \raisebox{-0.5mm}{\footnotesize{42}}}                	
                	\psfrag{60}{\hspace{-1mm} \raisebox{-0.5mm}{\footnotesize{60}}}
                 	\psfrag{1}{\hspace{-2mm} \raisebox{0mm}{\scriptsize{1}}}                	
                 	\psfrag{48}{\hspace{-2mm} \raisebox{0mm}{\footnotesize{48}}}
                 	\psfrag{12}{\hspace{-2mm} \raisebox{0mm}{\scriptsize{12}}}
                 	\psfrag{96}{\hspace{-2mm} \raisebox{0mm}{\footnotesize{96}}}                	
                 	\psfrag{24}{\hspace{-2mm} \raisebox{0mm}{\scriptsize{24}}}  
                     \includegraphics[height = 1.5in,width = 0.24\linewidth]{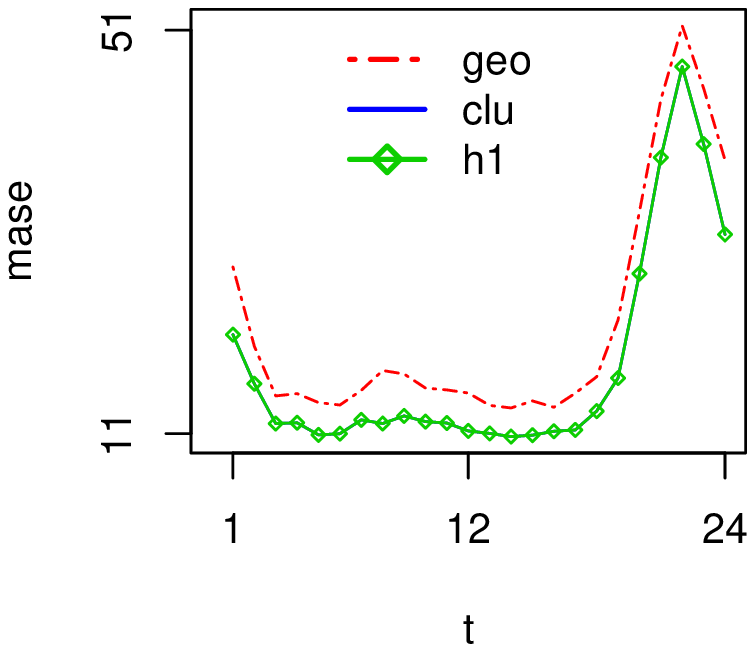}
                     \hfill%
                     \includegraphics[height = 1.5in,width = 0.24\linewidth]{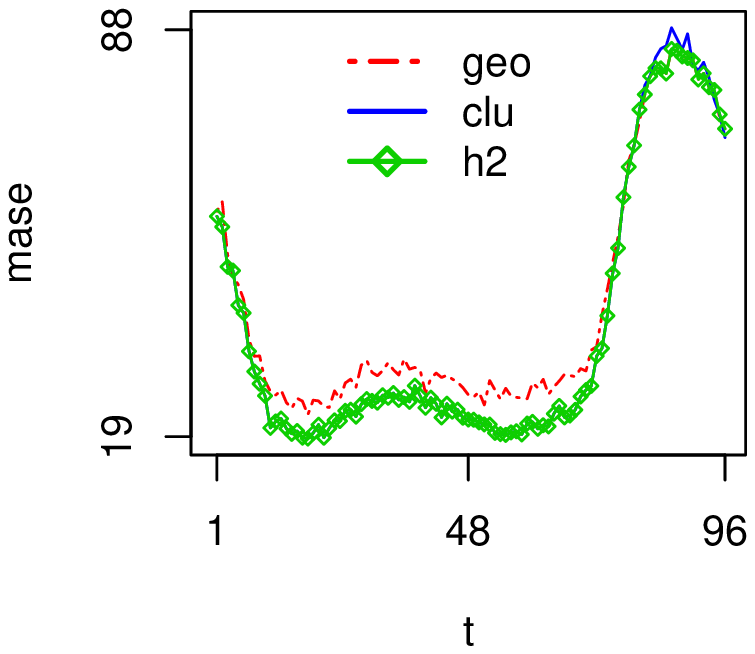}
                     \hfill%
                     \includegraphics[height = 1.5in,width = 0.24\linewidth]{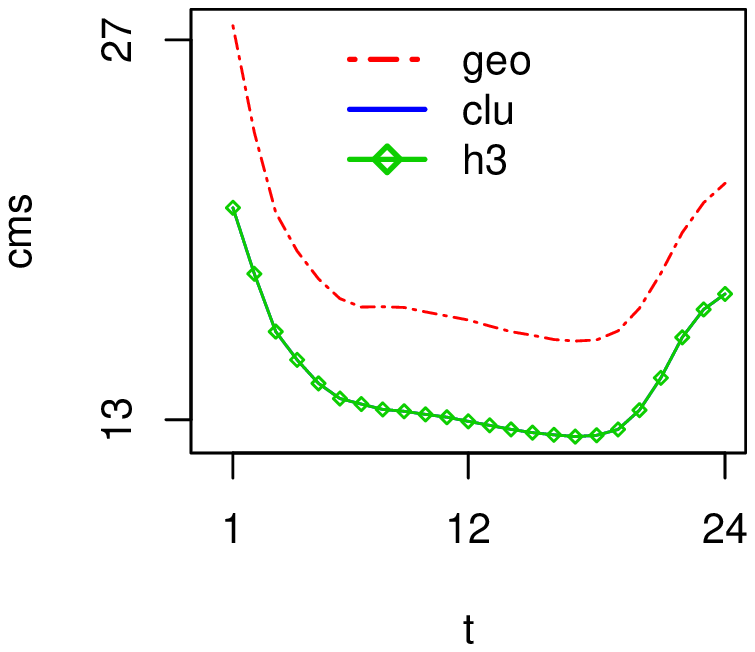}
                      \hfill%
                     \includegraphics[height = 1.5in,width = 0.24\linewidth]{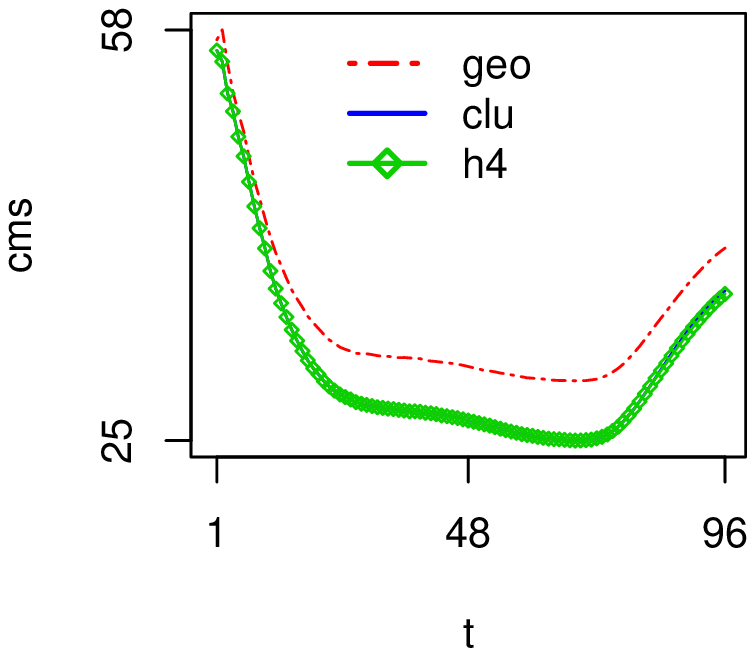}%
                   \vspace{-0.5cm}
                    \caption{Bengaluru data with SMAPE as the metric}
                    \label{smape_bang}
                \end{subfigure}%
                \vskip\baselineskip
                   \begin{subfigure}{\textwidth}
                     \psfrag{t}{\hspace{-7mm} \raisebox{2mm}{\footnotesize{Time step}}}
                	\psfrag{mase}{\hspace{-3mm}\raisebox{-1mm}{\scriptsize{MASE}}}
                	\psfrag{cms}{\hspace{-8mm}\raisebox{-1mm}{\footnotesize{Cum. Mean Error}}}
                     \psfrag{clu}{\hspace{-0.5mm}\raisebox{0mm}{\scriptsize{Voronoi}}}
                	\psfrag{geo}{\hspace{-0.5mm}\raisebox{0mm}{\scriptsize{Geohash}}}                	 
                	\psfrag{h1}{\hspace{-1.5mm}\raisebox{-2mm}{\shortstack{\scalebox{0.6}{dHEDGE}\\\scalebox{0.5}{(0.5,0.7)}}}}    
                	\psfrag{h2}{\hspace{0mm}\raisebox{-2mm}{\shortstack{\scalebox{0.6}{dHEDGE}\\\scalebox{0.5}{(0.1,0.3)}}}}  
                	\psfrag{h3}{\hspace{0mm}\raisebox{-2mm}{\shortstack{\scalebox{0.6}{dHEDGE}\\\scalebox{0.5}{(0.5,0.7)}}}}  
                	\psfrag{h4}{\hspace{0mm}\raisebox{-2mm}{\shortstack{\scalebox{0.6}{dHEDGE}\\\scalebox{0.5}{(0.1,0.3)}}}} 
                	\psfrag{1}{\hspace{-2mm} \raisebox{0mm}{\scriptsize{1}}}                	
                 	\psfrag{48}{\hspace{-2mm} \raisebox{0mm}{\footnotesize{48}}}
                 	\psfrag{12}{\hspace{-2mm} \raisebox{0mm}{\scriptsize{12}}}
                 	\psfrag{96}{\hspace{-2mm} \raisebox{0mm}{\footnotesize{96}}}                	
                 	\psfrag{24}{\hspace{-2mm} \raisebox{0mm}{\scriptsize{24}}} 
                	\psfrag{0.3}{\hspace{-1mm} \raisebox{-0.5mm}{\footnotesize{0.3}}}                	
                	\psfrag{1.4}{\hspace{-3mm} \raisebox{-0.5mm}{\footnotesize{1.4}}}
                	\psfrag{1.1}{\hspace{-3mm} \raisebox{-0.5mm}{\footnotesize{1.1}}} 	
                	\psfrag{0.4}{\hspace{-3mm} \raisebox{-0.5mm}{\footnotesize{0.4}}}                	
                	\psfrag{0.6}{\hspace{-1mm} \raisebox{-0.5mm}{\footnotesize{0.6}}}
                	\psfrag{1.0}{\hspace{-1mm} \raisebox{-0.5mm}{\footnotesize{1.0}}}
                	\psfrag{0.5}{\hspace{-1mm} \raisebox{-0.5mm}{\footnotesize{0.5}}}                	
                	\psfrag{0.7}{\hspace{-1mm} \raisebox{-0.5mm}{\footnotesize{0.7}}}
                	\psfrag{0.9}{\hspace{-3mm} \raisebox{-0.5mm}{\footnotesize{0.9}}}                	
                    \psfrag{0.30}{\hspace{-1mm} \raisebox{-0.5mm}{\footnotesize{0.3}}}                	
                	\psfrag{0.45}{\hspace{-1mm} \raisebox{-0.5mm}{\footnotesize{0.45}}}
                	\psfrag{0.60}{\hspace{-1mm} \raisebox{-0.5mm}{\footnotesize{0.6}}}        
                    \includegraphics[height = 1.5in,width = 0.24\linewidth]{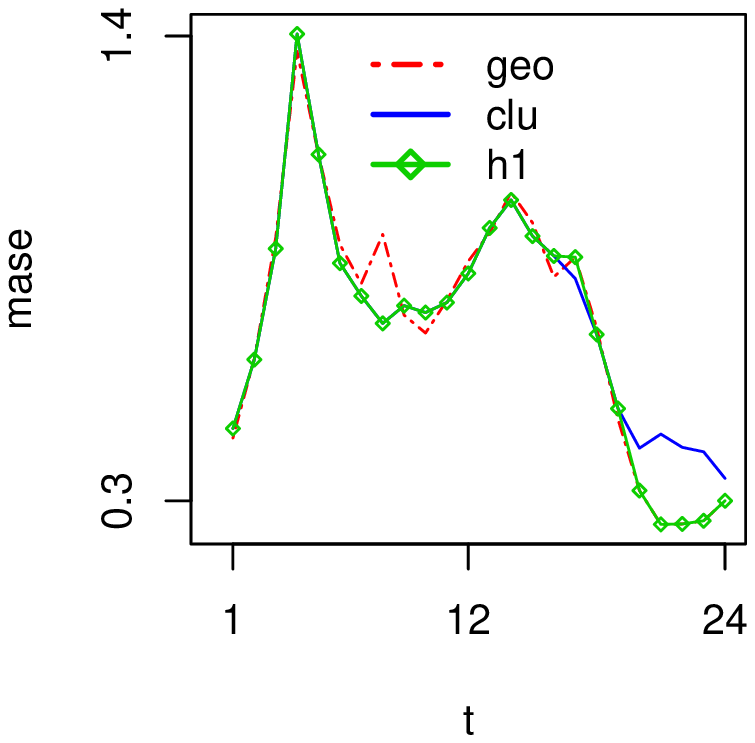}%
                       \hfill
                     \includegraphics[height = 1.5in,width = 0.24\linewidth]{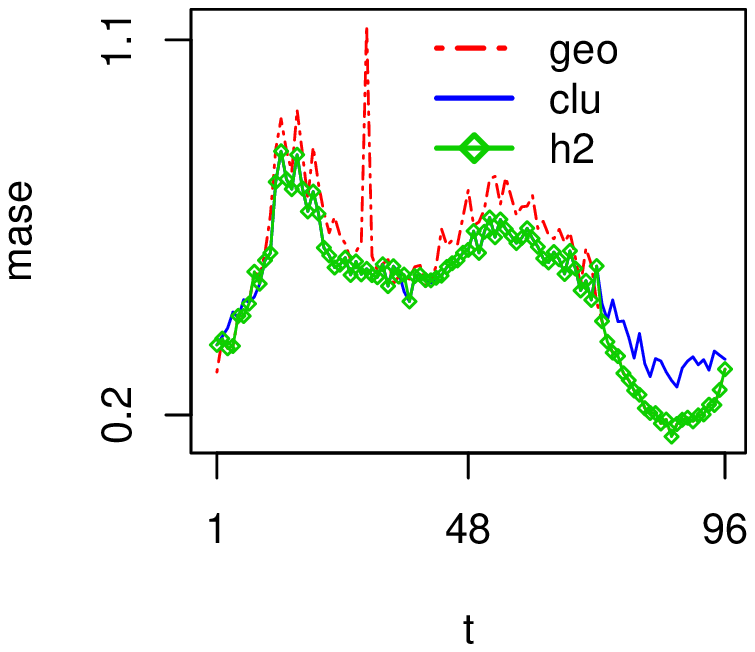}%
                     \hfill
                     \includegraphics[height = 1.5in,width = 0.24\linewidth]{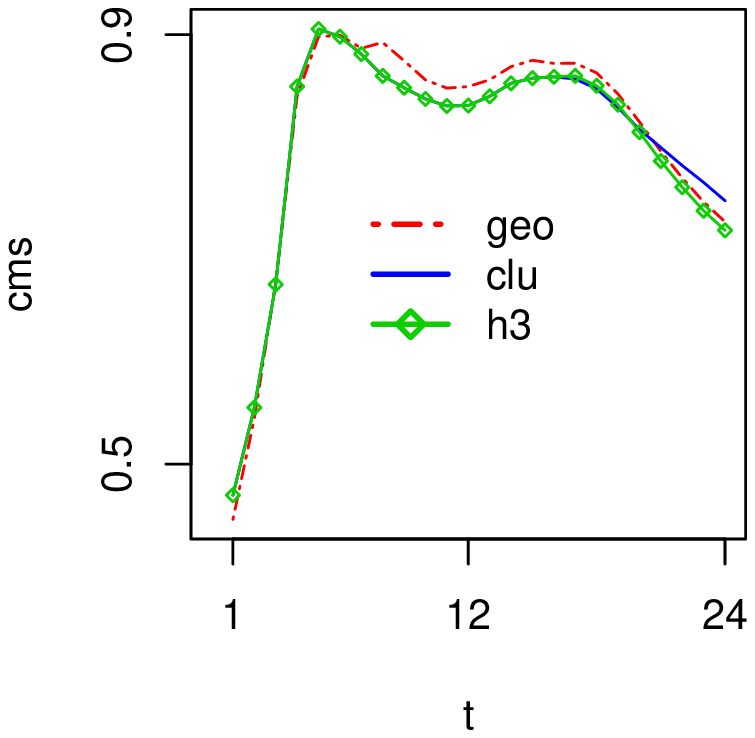}%
                       \hfill
                     \includegraphics[height = 1.5in,width = 0.24\linewidth]{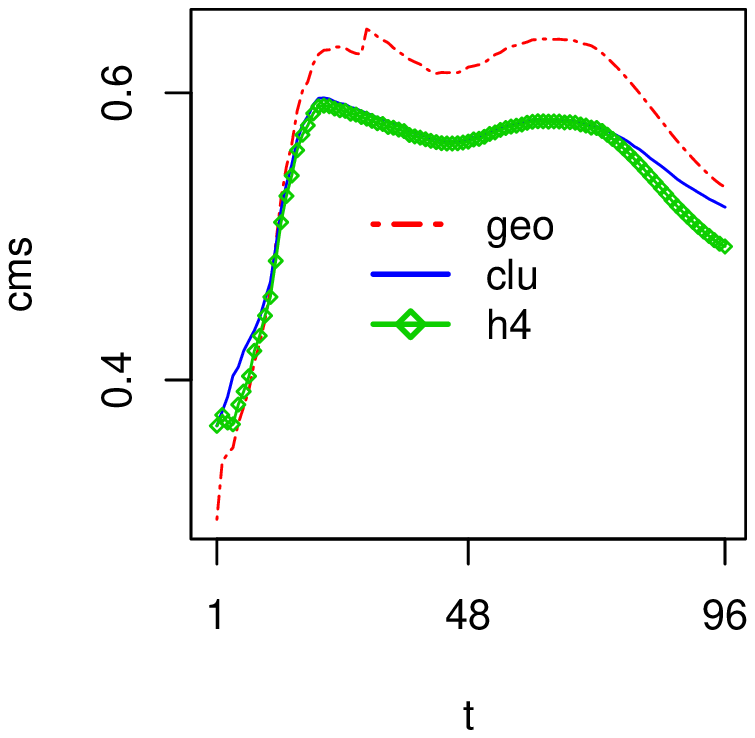}
                    \vspace{-0.5cm}
                    \caption{Bengaluru data with MASE as the metric}
                    \label{mape_bang}
                \end{subfigure}%
                \vskip\baselineskip
                \begin{subfigure}{\textwidth}
                     \psfrag{t}{\hspace{-7mm} \raisebox{2mm}{\footnotesize{Time step}}}
                	\psfrag{mase}{\hspace{-3mm}\raisebox{-1mm}{\scriptsize{SMAPE}}}
                	\psfrag{cms}{\hspace{-8mm}\raisebox{-1mm}{\footnotesize{Cum. Mean Error}}}
                     \psfrag{clu}{\hspace{0mm}\raisebox{0mm}{\scriptsize{Voronoi}}}
                	\psfrag{geo}{\hspace{0mm}\raisebox{0mm}{\scriptsize{Geohash}}}                	 
                	\psfrag{h1}{\hspace{0mm}\raisebox{-2mm}{\shortstack{\scalebox{0.6}{dHEDGE}\\\scalebox{0.5}{(0.1,0.1)}}}}      
                	\psfrag{h2}{\hspace{0mm}\raisebox{-2mm}{\shortstack{\scalebox{0.6}{dHEDGE}\\\scalebox{0.5}{(0.1,0.1)}}}}  
                	\psfrag{h3}{\hspace{0mm}\raisebox{-2mm}{\shortstack{\scalebox{0.6}{dHEDGE}\\\scalebox{0.5}{(0.1,0.1)}}}}    
                	\psfrag{h4}{\hspace{0mm}\raisebox{-2mm}{\shortstack{\scalebox{0.6}{dHEDGE}\\\scalebox{0.5}{(0.1,0.1)}}}}
                	\psfrag{1}{\hspace{-2mm} \raisebox{0mm}{\scriptsize{1}}}                	
                 	\psfrag{48}{\hspace{-2mm} \raisebox{0mm}{\footnotesize{48}}}
                 	\psfrag{12}{\hspace{-2mm} \raisebox{0mm}{\scriptsize{12}}}
                 	\psfrag{96}{\hspace{-2mm} \raisebox{0mm}{\footnotesize{96}}}                	
                 	\psfrag{24}{\hspace{-2mm} \raisebox{0mm}{\scriptsize{24}}}
                 	\psfrag{10}{\hspace{-1mm} \raisebox{-0.5mm}{\footnotesize{10}}}                	
                	\psfrag{35}{\hspace{-2mm} \raisebox{-0.5mm}{\footnotesize{35}}}
                	\psfrag{20}{\hspace{-1mm} \raisebox{-0.5mm}{\footnotesize{20}}} 	
                	\psfrag{16}{\hspace{-1mm} \raisebox{-0.5mm}{\footnotesize{16}}}                	
                	\psfrag{45}{\hspace{-1mm} \raisebox{-0.5mm}{\footnotesize{45}}}
                	\psfrag{71}{\hspace{-2mm} \raisebox{-0.5mm}{\footnotesize{71}}}
                	\psfrag{32}{\hspace{-1mm} \raisebox{-0.5mm}{\footnotesize{32}}}                	
                	\psfrag{54}{\hspace{-2mm} \raisebox{-0.5mm}{\footnotesize{54}}}
                	\psfrag{30}{\hspace{-1mm} \raisebox{-0.5mm}{\footnotesize{30}}}                	
                	\psfrag{42}{\hspace{-1mm} \raisebox{-0.5mm}{\footnotesize{42}}}
                	\psfrag{55}{\hspace{-1mm} \raisebox{-0.5mm}{\footnotesize{55}}}
                  	   \includegraphics[height = 1.5in,width = 0.24\linewidth]{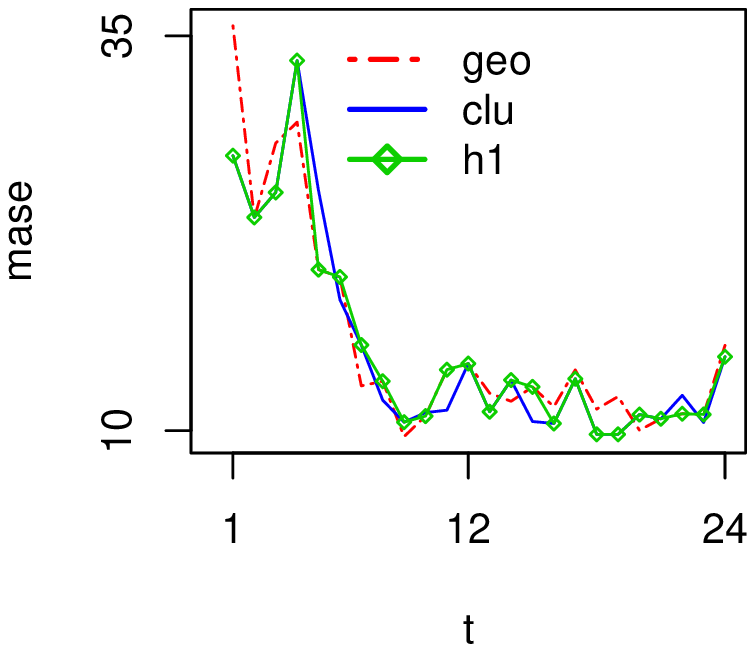}%
                     \hfill
                     \includegraphics[height = 1.5in,width = 0.24\linewidth]{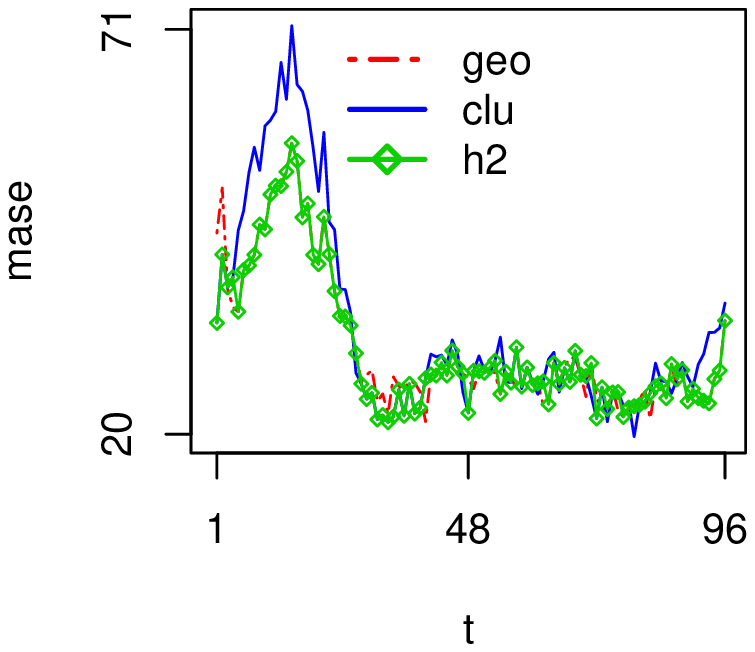}%
                     \hfill
                     \includegraphics[height = 1.5in,width = 0.24\linewidth]{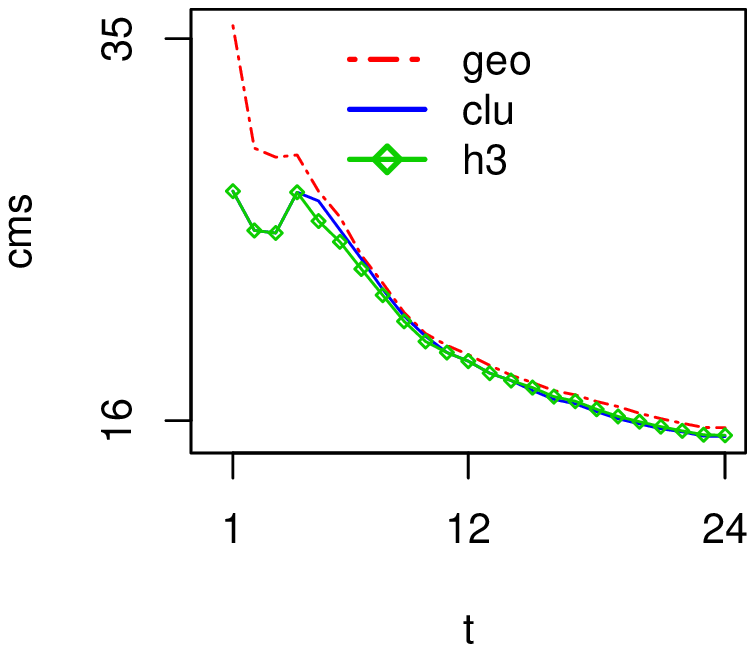}%
                       \hfill
                     \includegraphics[height = 1.5in,width = 0.24\linewidth]{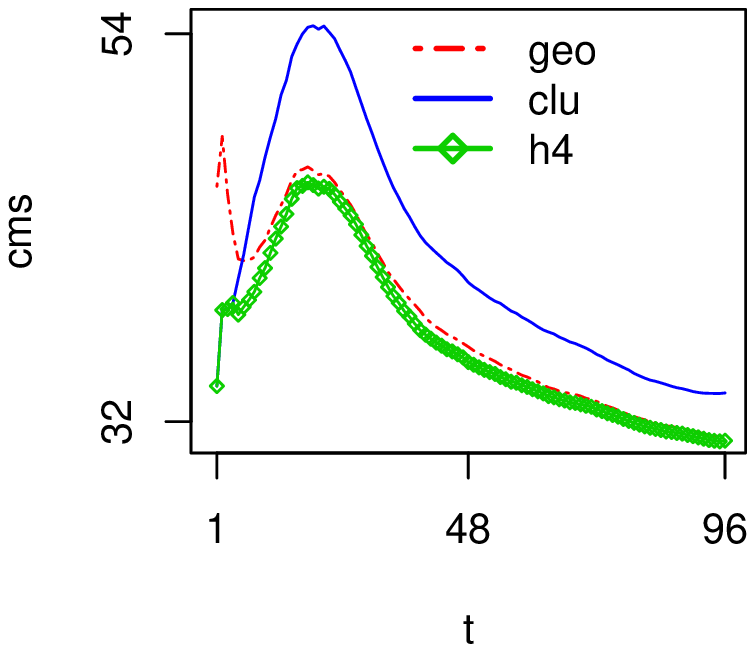}
                    \vspace{-0.5cm}
                     \caption{New York data with SMAPE as the metric}
                    \label{smape_nyc}
               \end{subfigure}%
                \vskip\baselineskip
                     \begin{subfigure}{\textwidth}
                     \psfrag{t}{\hspace{-7mm} \raisebox{2mm}{\footnotesize{Time step}}}
                	\psfrag{mase}{\hspace{-3mm}\raisebox{-1mm}{\scriptsize{MASE}}}
                	\psfrag{cms}{\hspace{-8mm}\raisebox{-1mm}{\footnotesize{Cum. Mean Error}}}
                     \psfrag{clu}{\hspace{0mm}\raisebox{0mm}{\scriptsize{Voronoi}}}
                	\psfrag{geo}{\hspace{0mm}\raisebox{0mm}{\scriptsize{Geohash}}}                	 
                	\psfrag{h1}{\hspace{0mm}\raisebox{-2mm}{\shortstack{\scalebox{0.6}{dHEDGE}\\\scalebox{0.5}{(0.1,0.1)}}}}    
                	\psfrag{h2}{\hspace{0mm}\raisebox{-2mm}{\shortstack{\scalebox{0.6}{dHEDGE}\\\scalebox{0.5}{(0.1,0.5)}}}}  
                	\psfrag{h3}{\hspace{0mm}\raisebox{-2mm}{\shortstack{\scalebox{0.6}{dHEDGE}\\\scalebox{0.5}{(0.1,0.1)}}}}  
                	\psfrag{h4}{\hspace{0mm}\raisebox{-2mm}{\shortstack{\scalebox{0.6}{dHEDGE}\\\scalebox{0.5}{(0.1,0.5)}}}} 
                	\psfrag{1}{\hspace{-2mm} \raisebox{0mm}{\scriptsize{1}}}                	
                 	\psfrag{48}{\hspace{-2mm} \raisebox{0mm}{\footnotesize{48}}}
                 	\psfrag{12}{\hspace{-2mm} \raisebox{0mm}{\scriptsize{12}}}
                 	\psfrag{96}{\hspace{-2mm} \raisebox{0mm}{\footnotesize{96}}}                	
                 	\psfrag{24}{\hspace{-2mm} \raisebox{0mm}{\scriptsize{24}}} 
                    \psfrag{0.3}{\hspace{-1mm} \raisebox{-0.5mm}{\footnotesize{0.3}}}                	
                	\psfrag{0.8}{\hspace{-1mm} \raisebox{-0.5mm}{\footnotesize{0.8}}}
                	\psfrag{1.3}{\hspace{-1mm} \raisebox{-0.5mm}{\footnotesize{1.3}}} 	
                	\psfrag{0.2}{\hspace{-1mm} \raisebox{-0.5mm}{\footnotesize{0.2}}}                	
                	\psfrag{0.4}{\hspace{-1mm} \raisebox{-0.5mm}{\footnotesize{0.4}}}
                	\psfrag{1.6}{\hspace{-2mm} \raisebox{-0.5mm}{\footnotesize{1.6}}}
                	\psfrag{0.5}{\hspace{-1mm} \raisebox{-0.5mm}{\footnotesize{0.5}}}                	
                	\psfrag{1.4}{\hspace{-2mm} \raisebox{-0.5mm}{\footnotesize{1.4}}}
                	\psfrag{1.2}{\hspace{-1mm} \raisebox{-0.5mm}{\footnotesize{1.2}}}
                	\psfrag{0.9}{\hspace{-1mm} \raisebox{-0.5mm}{\footnotesize{0.9}}}                	
                  	\includegraphics[height = 1.5in,width = 0.24\linewidth]{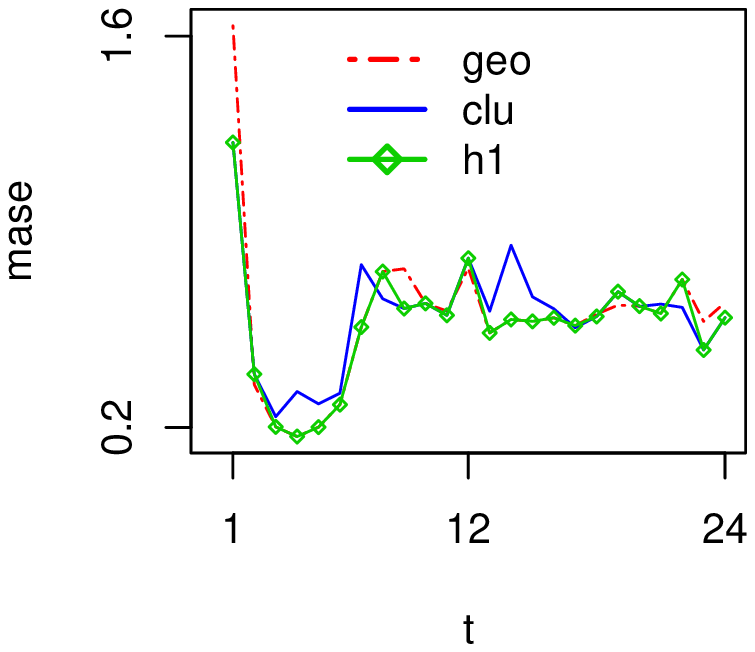}%
                       \hfill
                     \includegraphics[height = 1.5in,width = 0.24\linewidth]{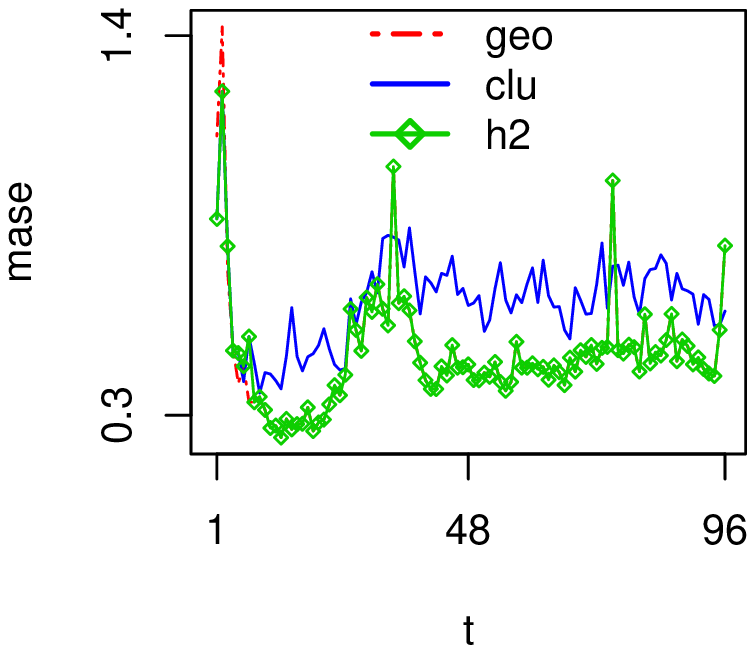}%
                     \hfill
                     \includegraphics[height = 1.5in,width = 0.24\linewidth]{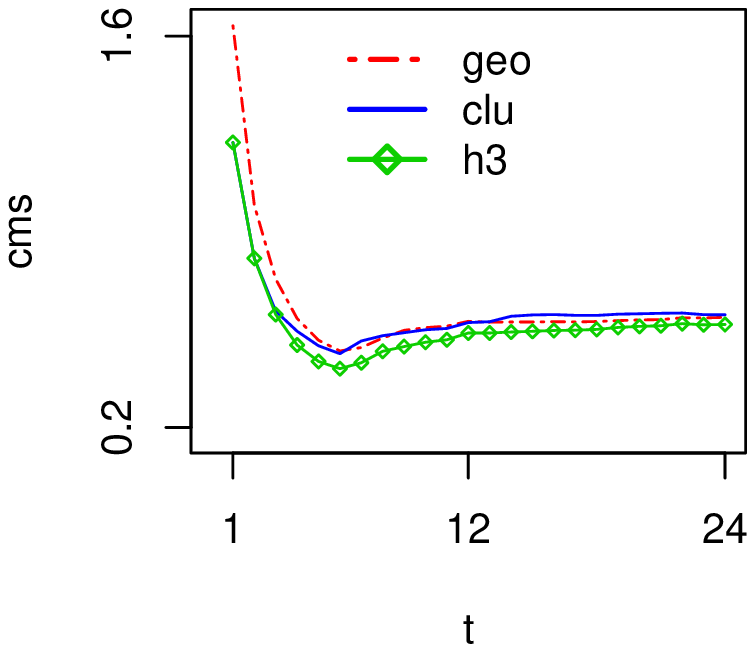}%
                       \hfill
                     \includegraphics[height = 1.5in,width = 0.24\linewidth]{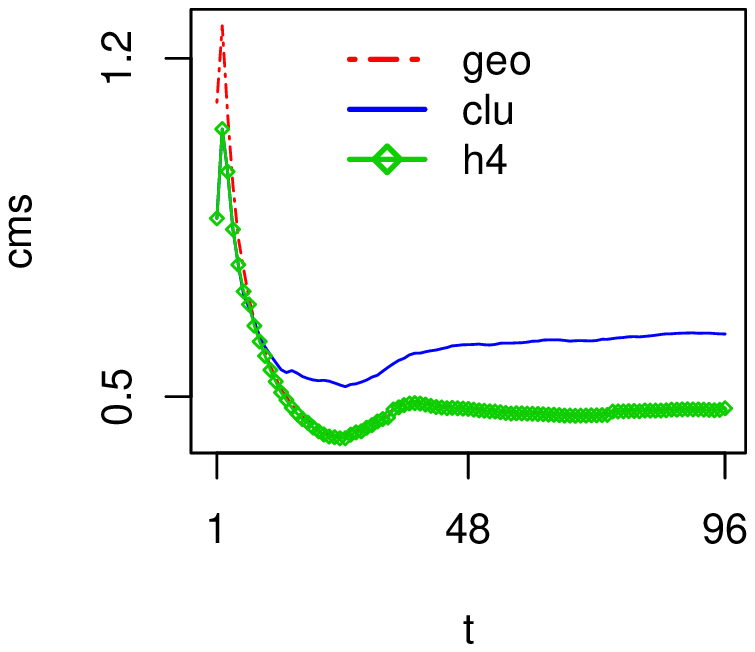}
                     \vspace{-0.5cm}
                    \caption{New York data with MASE as the metric}
                    \label{mase_nyc}
                \end{subfigure}%
                  \vskip\baselineskip
                     \begin{subfigure}{\textwidth}
                     \psfrag{t}{\hspace{-7mm} \raisebox{2mm}{\footnotesize{Time step}}}
                	\psfrag{mase}{\hspace{-3mm}\raisebox{-1mm}{\scriptsize{MASE}}}
                	\psfrag{cms1}{\hspace{-9mm}\raisebox{-1mm}{\footnotesize{Cum. mean SMAPE}}}		
                	\psfrag{cms2}{\hspace{-9mm}\raisebox{-1mm}{\footnotesize{Cum. mean MASE}}}	
                     \psfrag{clu}{\hspace{0mm}\raisebox{-0mm}{\scriptsize{Voronoi}}}
                	\psfrag{geo}{\hspace{0mm}\raisebox{-0mm}{\scriptsize{Geohash}}}                	 
                	\psfrag{h1}{\hspace{0mm}\raisebox{-2mm}{\shortstack{\scalebox{0.6}{dHEDGE}\\\scalebox{0.5}{(0.1,0.3)}}}}    
                	\psfrag{h2}{\hspace{0mm}\raisebox{-2mm}{\shortstack{\scalebox{0.6}{dHEDGE}\\\scalebox{0.5}{(0.1,0.7)}}}}  
                	\psfrag{h4}{\hspace{0mm}\raisebox{-2mm}{\shortstack{\scalebox{0.6}{dHEDGE}\\\scalebox{0.5}{(0.1,0.8)}}}}   
                	\psfrag{h3}{\hspace{0mm}\raisebox{-2mm}{\shortstack{\scalebox{0.6}{dHEDGE}\\\scalebox{0.5}{(0.1,0.1)}}}} 
                	\psfrag{1}{\hspace{-2mm} \raisebox{0mm}{\scriptsize{1}}}                	
                 	\psfrag{48}{\hspace{-2mm} \raisebox{0mm}{\footnotesize{48}}}
                 	\psfrag{12}{\hspace{-2mm} \raisebox{0mm}{\scriptsize{12}}}
                 	\psfrag{96}{\hspace{-2mm} \raisebox{0mm}{\footnotesize{96}}}                	
                 	\psfrag{24}{\hspace{-2mm} \raisebox{0mm}{\scriptsize{24}}} 
                    \psfrag{14}{\hspace{-1mm} \raisebox{-0.5mm}{\footnotesize{14}}}                	
                	\psfrag{39}{\hspace{-2mm} \raisebox{-0.5mm}{\footnotesize{39}}}
                	\psfrag{40}{\hspace{-1mm} \raisebox{-0.5mm}{\footnotesize{40}}} 	
                	\psfrag{0.2}{\hspace{-1mm} \raisebox{-0.5mm}{\footnotesize{0.2}}}                	
                	\psfrag{0.7}{\hspace{-1mm} \raisebox{-0.5mm}{\footnotesize{0.7}}}
                	\psfrag{0.90}{\hspace{-1mm} \raisebox{-0.5mm}{\footnotesize{0.9}}}
                	\psfrag{72}{\hspace{-1mm} \raisebox{-0.5mm}{\footnotesize{72}}}                	
                	\psfrag{95}{\hspace{-1mm} \raisebox{-0.5mm}{\footnotesize{95}}}
                	\psfrag{128}{\hspace{-3mm} \raisebox{-0.5mm}{\footnotesize{128}}}
                	\psfrag{0.6}{\hspace{-1mm} \raisebox{-0.5mm}{\footnotesize{0.6}}}             
                	\psfrag{1.0}{\hspace{-1mm} \raisebox{-0.5mm}{\footnotesize{1}}}                	
                	\psfrag{1.4}{\hspace{-1mm} \raisebox{-0.5mm}{\footnotesize{1.4}}}                	

                  	\includegraphics[height = 1.5in,width = 0.24\linewidth]{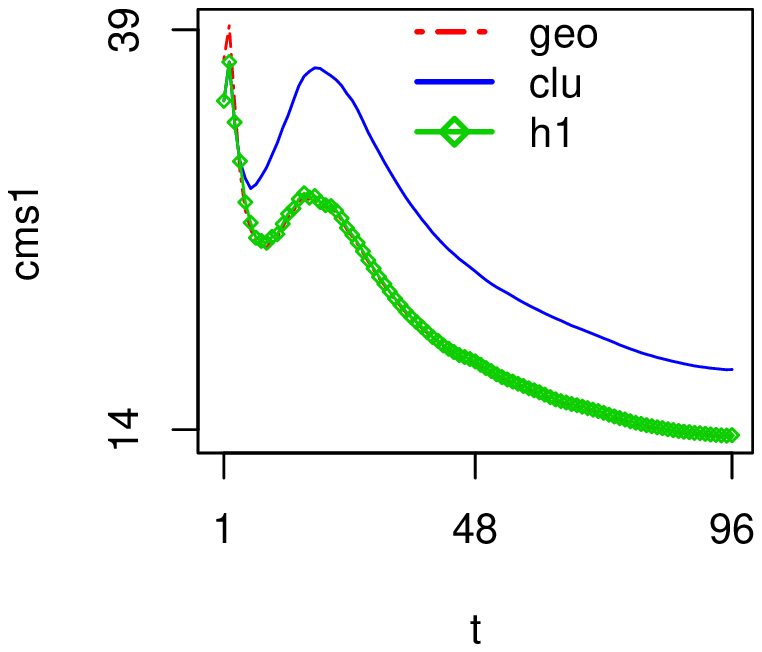}%
                       \hfill
                     \includegraphics[height = 1.5in,width = 0.24\linewidth]{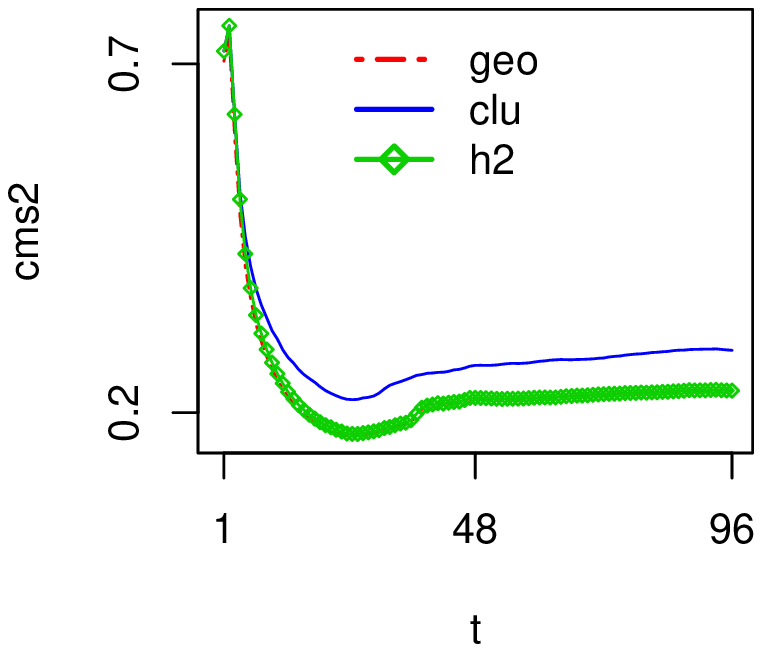}%
                     \hfill
                     \includegraphics[height = 1.5in,width = 0.24\linewidth]{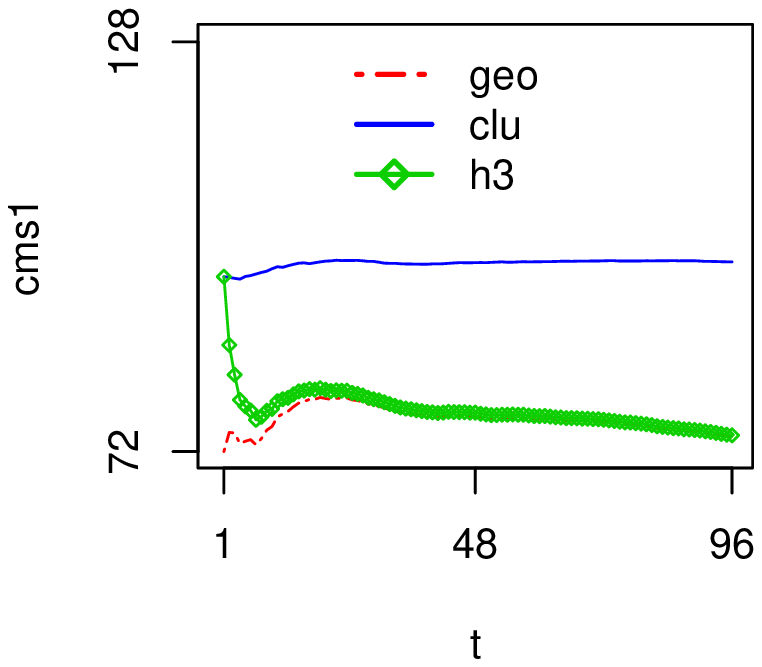}%
                       \hfill
                     \includegraphics[height = 1.5in,width = 0.24\linewidth]{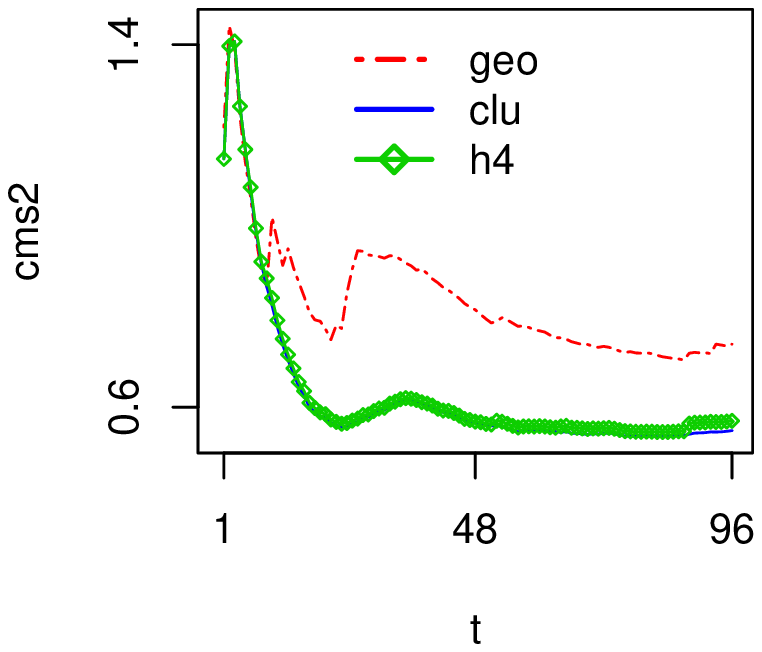}
                     \vspace{-0.5cm}
                    \caption{Manhattan-Bronx and Brooklyn-Queens data sets with SMAPE and MASE as metrics}
                    \label{nyc_boroughs}
                \end{subfigure}%
                \vspace{0.4cm}
                 \caption{The prediction performance of dHEDGE ($\beta$,$\gamma$) algorithm based model against single strategy based models at 15 and 60 minute aggregation levels for the two cities with SMAPE and MASE as metrics are plotted in the first two sub figures of (a)-(d). Cumulative mean error performance of models based on Geohash, Voronoi and hybrid strategies are provided in the last two subfigures of (a)-(d). The first two and the last two subfigures of (e) corresponds to cumulative mean error performances on Manhattan-Bronx and Brooklyn-Queens boroughs respectively at 15 minute aggregation levels.}
                 \label{hedgeall}
\end{figure*}

The weights can be initialized either uniformly or based on some prior knowledge about the experts. The factor $\gamma$ gives more leverage to observations made in the recent past compared to the observations in the distant past. Thus, the expert which have been performing well in the recent past is boosted. By tuning the parameter $\gamma$, we can manipulate the dHEDGE algorithm to respond to sudden changes in the average error performance. By setting $\gamma$ = 1, this algorithm becomes HEDGE with no forgetting factor. When modified dHEDGE is used, we observe that the algorithm tends to choose the best expert quite rapidly compared to the original HEDGE (Figure \ref{after_hedge_disc}).
\begin{table*}[h!]
\centering
\scalebox{0.7}{
\begin{tabular}{|c||c||c|c|c||c|c|c||c|c|c||c|c|c|}
\hline
\multirow{3}{*}{\textbf{Data set}} & \multirow{3}{*}{\textbf{Tessellation size}}                                                                & \multicolumn{12}{c|}{\textbf{Sampling period}}                                                                                                                                                                                                                                                                                    \\ \cline{3-14} 
                                   &                                                                                                            & \multicolumn{3}{c||}{5}                                                         & \multicolumn{3}{c||}{15}                                                        & \multicolumn{3}{c||}{30}                                                        & \multicolumn{3}{c|}{60}                                                        \\ \cline{3-14} 
                                   &                                                                                                            & Vor                   & Geo                   & dHEDGE                         & Vor                   & Geo                   & dHEDGE                         & Vor                   & Geo                   & dHEDGE                         & Vor                   & Geo                   & dHEDGE                         \\ \hline \hline
\multirow{6}{*}{Bengaluru}         & \multirow{2}{*}{\begin{tabular}[c]{@{}c@{}}6-level geohash\\ \& K= 740 \\ \end{tabular}} & \multirow{2}{*}{63.2} & \multirow{2}{*}{60.8} & \textbf{60.7}                  & \multirow{2}{*}{37.1} & \multirow{2}{*}{40.4} & \textbf{36.5}                  & \multirow{2}{*}{25.6} & \multirow{2}{*}{29.6} & \textbf{25.6}                  & \multirow{2}{*}{17.6} & \multirow{2}{*}{21.7} & \textbf{17.6}                  \\ \cline{5-5} \cline{8-8} \cline{11-11} \cline{14-14} 
                                   &                                                                                                            &                       &                       & (0.1,0.4)                      &                       &                       & (0.1,0.4)                      &                       &                       & (0.1,0.1)                      &                       &                       & (0.1,0.1)                      \\ \cline{2-14} 
                                   & \multirow{2}{*}{\begin{tabular}[c]{@{}c@{}}sub 5-level geohash\\ \& K= 114\\ \end{tabular}}                         & \multirow{2}{*}{23}   & \multirow{2}{*}{20.7} & \textbf{20.5}                  & \multirow{2}{*}{12.8} & \multirow{2}{*}{19.3} & \textbf{12.6}                  & \multirow{2}{*}{9.7}  & \multirow{2}{*}{16.7} & \textbf{9.6}                   & \multirow{2}{*}{6.9}  & \multirow{2}{*}{12.8} & \textbf{7.0}                   \\ \cline{5-5} \cline{8-8} \cline{11-11} \cline{14-14} 
                                   &                                                                                                            &                       &                       & (0.1,0.1)                      &                       &                       & (0.1,0.6)                      &                       &                       & (0.1,0.1)                      &                       &                       & (0.1,0.3)                      \\ \cline{2-14} 
                                   & \multirow{2}{*}{\begin{tabular}[c]{@{}c@{}}5-level geohash\\ \& K = 30\\\end{tabular}} & \multirow{2}{*}{10.9} & \multirow{2}{*}{10.5} & \textbf{9.7}                   & \multirow{2}{*}{7.4}  & \multirow{2}{*}{9.4}  & \textbf{7.3}                   & \multirow{2}{*}{6.4}  & \multirow{2}{*}{7.4}  & \textbf{6.4}                   & \multirow{2}{*}{5.4}  & \multirow{2}{*}{5.8}  & \textbf{5.4}                   \\ \cline{5-5} \cline{8-8} \cline{11-11} \cline{14-14} 
                                   &                                                                                                            &                       &                       & (0.1,0.9)                      &                       &                       & (0.1,0.7)                      &                       &                       & 0.9,0.7                        &                       &                       & 0.9,0.9                        \\ \hline
\multirow{6}{*}{New York}          & \multirow{2}{*}{\begin{tabular}[c]{@{}c@{}}6-level geohash\\ \& K= 780\end{tabular}}                     & \multirow{2}{*}{61.4} & \multirow{2}{*}{40.0} & \textbf{40.0}                  & \multirow{2}{*} {33.6}             & \multirow{2}{*} {31.2} & \textbf{31}                    & \multirow{2}{*}{24.8} & \multirow{2}{*}{24.4} & \textbf{23.8}                  & \multirow{2}{*}{15.2} & \multirow{2}{*}{15.6} & \textbf{15.2}                  \\ \cline{5-5} \cline{8-8} \cline{11-11} \cline{14-14} 
                                   &                                                                                                            &                       &                       & (0.1,0.1)                      &                       &                       & (0.1,0.1)                      &                       &                       & (0.1,0.6)                      &                       &                       & (0.1,0.1)                      \\ \cline{2-14} 
                                   & \multirow{2}{*}{\begin{tabular}[c]{@{}c@{}}sub 5-level geohash\\ \& K= 120\end{tabular}}          & \multirow{2}{*}{23.7} & \multirow{2}{*}{15.9} & \textbf{15.5}                  & \multirow{2}{*}{13.9} & \multirow{2}{*}{12.4} & \textbf{12.1}                  & \multirow{2}{*}{11}   & \multirow{2}{*}{11.6} & \textbf{10.6}                  & \multirow{2}{*}{9.4}  & \multirow{2}{*}{11.9} & \textbf{9.3}                   \\ \cline{5-5} \cline{8-8} \cline{11-11} \cline{14-14} 
                                   &                                                                                                            &                       &                       & (0.8,0.8)                      &                       &                       & (0.7,0.2)                      &                       &                       & (0.1,0.3)                      &                       &                       & (0.1,0.4)                      \\ \cline{2-14} 
                                   & \multirow{2}{*}{\begin{tabular}[c]{@{}c@{}}5-level geohash\\ \& K= 31\end{tabular}}                      & \multirow{2}{*}{10.9} & \multirow{2}{*}{10.5} & \textbf{10.4}                  & \multirow{2}{*}{9.7}  & \multirow{2}{*}{8.7}  & \textbf{8.3}                   & \multirow{2}{*}{11.2} & \multirow{2}{*}{7.7}  & \textbf{7.9}                   & \multirow{2}{*}{10.6}  & \multirow{2}{*}{12.8}  & \textbf{10.8}                     \\ \cline{5-5} \cline{8-8} \cline{11-11} \cline{14-14} 
                                   &                                                                                                            &                       &                       & (0.1,0.9)                      &                       &                       & (0.1,0.7)                      &                       &                       & (0.1,0.1)                      &                       &                       & (0.1,0.8)                      \\ \hline
\multirow{6}{*}{Manhattan-Bronx}   & \multirow{2}{*}{\begin{tabular}[c]{@{}c@{}}6-level geohash\\ \& K= 169\end{tabular}}                     & \multirow{2}{*}{29.9} & \multirow{2}{*}{20.6} & \textbf{20.6}                  & \multirow{2}{*}{17}   & \multirow{2}{*}{16.5} & \textbf{16.5}                  & \multirow{2}{*}{13.2} & \multirow{2}{*}{11.1} & \textbf{11.1}                  & \multirow{2}{*}{9.15} & \multirow{2}{*}{10.5} & \textbf{9.15}                  \\ \cline{5-5} \cline{8-8} \cline{11-11} \cline{14-14} 
                                   &                                                                                                            &                       &                       & (0.1,0.2)                      &                       &                       & (0.1,0.3)                      &                       &                       & (0.1,0.1)                      &                       &                       & (0.1,0.1)                      \\ \cline{2-14} 
                                   & \multirow{2}{*}{\begin{tabular}[c]{@{}c@{}}sub 5-level geohash\\ \& K= 26\end{tabular}}             & \multirow{2}{*}{11.6} & \multirow{2}{*}{14.7} & \textbf{11.4}                  & \multirow{2}{*}{8.3}  & \multirow{2}{*}{12.9} & \textbf{8.4}                   & \multirow{2}{*}{7.4}  & \multirow{2}{*}{10.2} & \textbf{7.5}                   & \multirow{2}{*}{6.8}  & \multirow{2}{*}{9.3}  & \textbf{6.9}                   \\ \cline{5-5} \cline{8-8} \cline{11-11} \cline{14-14} 
                                   &                                                                                                            &                       &                       & (0.1,0.8)                      &                       &                       & (0.1,0.3)                      &                       &                       & (0.1,0.2)                      &                       &                       & (0.1,0.1)                      \\ \cline{2-14} 
                                   & \multirow{2}{*}{\begin{tabular}[c]{@{}c@{}}5-level geohash\\ \& K = 7\end{tabular}}                      & \multirow{2}{*}{7.9}  & \multirow{2}{*}{7.2}  & \textbf{6.9}                   & \multirow{2}{*}{6.2}  & \multirow{2}{*}{6.7}  & \textbf{6.2}                   & \multirow{2}{*}{5.5}  & \multirow{2}{*}{5.7}  & \textbf{5.1}                   & \multirow{2}{*}{5.9}  & \multirow{2}{*}{5.4}  & \textbf{5.5}                   \\ \cline{5-5} \cline{8-8} \cline{11-11} \cline{14-14} 
                                   &                                                                                                            &                       &                       & (0.1,0.7)                      &                       &                       & (0.7,0.9)                      &                       &                       & (0.7,0.2)                      &                       &                       & (0.6,0.9)                      \\ \hline
\multirow{6}{*}{Brooklyn-Queens}   & \multirow{2}{*}{\begin{tabular}[c]{@{}c@{}}6-level geohash\\ \& K = 460\end{tabular}}                      & \multirow{2}{*}{90.2} & \multirow{2}{*}{57.2} & \textbf{57.2}                  & \multirow{2}{*}{97.6} & \multirow{2}{*}{74.9} & \textbf{74.2}                  & \multirow{2}{*}{73.9} & \multirow{2}{*}{41.8} & \textbf{41.8}                  & \multirow{2}{*}{62.6} & \multirow{2}{*}{52.9} & \textbf{52.9}                  \\ \cline{5-5} \cline{8-8} \cline{11-11} \cline{14-14} 
                                   &                                                                                                            &                       &                       & (0.1,0.1)                      &                       &                       & (0.1,0.1)                      &                       &                       & (0.1,0.1)                      &                       &                       & (0.1,0.1)                      \\ \cline{2-14} 
                                   & \multirow{2}{*}{\begin{tabular}[c]{@{}c@{}}sub 5-level geohash\\ \& K= 71\end{tabular}}            & \multirow{2}{*}{73.0} & \multirow{2}{*}{35.1} & \textbf{35.1}                  & \multirow{2}{*}{49.3} & \multirow{2}{*}{27.2} & \textbf{27.2}                  & \multirow{2}{*}{37}   & \multirow{2}{*}{22.9} & \textbf{22.9}                  & \multirow{2}{*}{26.2} & \multirow{2}{*}{15.9} & \textbf{15.8}                  \\ \cline{5-5} \cline{8-8} \cline{11-11} \cline{14-14} 
                                   &                                                                                                            &                       &                       & (0.1,0.1)                      &                       &                       & (0.1,0.1)                      &                       &                       & (0.1,0.4)                      &                       &                       & (0.1,0.3)                      \\ \cline{2-14} 
                                   & \multirow{2}{*}{\begin{tabular}[c]{@{}c@{}}5-level geohash\\ \& K = 18\end{tabular}}                       & \multirow{2}{*}{42.4} & \multirow{2}{*}{19.5} & \textbf{19.5}                  & \multirow{2}{*}{23.1} & \multirow{2}{*}{28.8} & \textbf{23.1}                  & \multirow{2}{*}{18.2} & \multirow{2}{*}{22.8} & \textbf{18.2}                  & \multirow{2}{*}{12.9} & \multirow{2}{*}{18}   & \textbf{12.9}                  \\ \cline{5-5} \cline{8-8} \cline{11-11} \cline{14-14} 
                                   &                                                                                                            &                       &                       & \multicolumn{1}{l||}{(0.1,0.4)} &                       &                       & \multicolumn{1}{l||}{(0.1,0.8)} &                       &                       & \multicolumn{1}{l||}{(0.1,0.8)} &                       &                       & \multicolumn{1}{l|}{(0.9,0.9)} \\ \hline
\end{tabular}}
\caption{The table contains the performance comparison of Voronoi, Geohash and dHEDGE using SMAPE, for the entire set of simulations done. We observe that dHEDGE performs superiorly for all the tested combinations on the 4 data sets. The unique set of parameters ($\beta$,$\gamma$) obtained for each test case are mentioned below the errors encountered using dHEDGE algorithm.  }
\label{allresults}
\end{table*}

The modified dHEDGE algorithm was run on both datasets at both sampling periods. The parameters $\beta$ and $\gamma$ for each scenario are chosen based on an independent validation set spanning over 24 hours. The performance of the algorithm can be seen in the Figure \ref{hedgeall}. We can see that the algorithm picks up the best shifting expert, by giving more weightage to the behavior of that expert in the recent past. Note that in Figure \ref{smape_bang}, the plots of Voronoi and dHEDGE overlap as Voronoi is consistently the best performer throughout the forecasting horizon. The last two sub figures in each of the Figures \ref{smape_bang}-\ref{mase_nyc} show the cumulative mean error behaviors of the strategies. The cumulative mean error plots reveal that the modified dHEDGE performs at least as good as the best expert. Whenever the experts in consideration have similar cumulative errors, the hybrid strategy has an error lower than both experts; else, the hybrid strategy consistently has a cumulative error equal to the best performing expert in the recent past. 
We also analyze the performance of the strategies on various boroughs of New York city, to avoid the issue of inter-island Voronoi tessellations. To further demonstrate the flexibility of our algorithm in adapting to various scenarios, we have repeated the simulations for temporal aggregation levels of 5 and 30 minutes. Multiple spatial resolutions are also explored to validate the applicability of our algorithm.  Irrespective of all these nuances, when dHEDGE was applied, the hybrid strategy still turned out to be the winner for all the datasets considered, with different metrics at multiple temporal and spatial scales. Table \ref{allresults} summarizes the performances of the strategies on the datasets, with SMAPE as the metric. Similar results was observed using MASE, and hence,are not included. We observe that the modified dHEDGE improves the prediction accuracy by picking the best strategy for that particular instant, for that city, at that temporal and spatial resolution. Please note that our policy does not switch continuously between strategies every $t$ minutes. This is because the dHEDGE chooses the winner strategy for an instant based on neither the current nor the immediate past performance, but on the \emph{recent} past performance of both experts. The dHEDGE switches only if one strategy performs better than the other for a \emph{period of time} in the past. On tracking the performance of our strategy for all the test cases mentioned in Table \ref{allresults}, we observe an average of only 1.5 switches per day at 60 min. aggregation levels. In other words, for 24 time steps, the hybrid strategy made less than 2 switches on average for all tested scenarios. For 30 min. (\emph{i.e.,} 48 steps) and 15 min. (\emph{i.e.,} 96 steps) aggregation levels, the average switches made were only 3.2 and 8.4 per day. With finer temporal resolutions, the number of switches increases, which is inevitable, considering the increase in the number of time steps in the forecast horizon. We feel that the observed number of switches is admissible, in return for a significant improvement in accuracy.

Therefore, based on our observations from two independent datasets, we find that there is no universal winning tessellation technique that works for all datasets. However, a hybrid technique can be developed for enhanced prediction performances on a broad range of datasets.  
\section{Conclusions}\label{conclusions}
Efficient spatial partitioning is a key step towards better location-based demand modeling and forecasting. To that end, we performed a comparison of two widely used spatial partitioning techniques, Voronoi tessellation and Geohashing. 
The study was conducted using two distinct datasets; mobile application log-in based taxi demand data set from Bengaluru, India and government-run street hailing Yellow taxi services data set from New York, USA.  A K-Means clustering algorithm was employed to generate demand clusters, that act as generating sites for the tessellations. 
In order to compare Voronoi based models and Geohash based models at multiple time scales, time-series techniques were applied to the datasets at temporal aggregation levels of 15 and 60 minutes. STL decomposition with ARIMA, ETS, and TBATS were shortlisted as the suitable models. We noticed that neither Geohash nor Voronoi tessellation technique individually proved to be optimal for the entire forecast horizon. Additionally, models based on Voronoi had a superior performance over models based on Geohash when the demand density is low, and vice versa. While Voronoi tessellation appears to be the recommended strategy for tessellating Bengaluru, Geohash tessellation was the winning strategy for the New York city. Thus, we conclude that the tessellation strategy is heavily dependent on the demand density in each partition, and the geography of the city. 

The lack of a clear winning strategy prompted us to devise a hybrid tessellation strategy by combining models based on the well known HEDGE algorithm. We modified dHEDGE, which is a discounted version of the HEDGE algorithm, to suit our application. Our algorithm is shown to always pick up the best possible strategy at each time step in the forecast horizon. The tessellation performance is no longer affected by the time of the day, or the features of the underlying data set. Our hybrid tessellation strategy was clearly the winning strategy for all the datasets considered, performing consistently better at multiple time scales with different performance metrics. 

 This work is directed towards a comparison of tessellation strategies, where we have focused on combining the strategies. More tessellation experts can be added to check the robustness of our dHEDGE algorithm. Recent developments in Recurrent Neural Networks have shown promises in the prediction domain \cite{xu2017real}, which can be utilized to improve predictions. As part of future works, non-parametric techniques such as Support Vector Regression, Artificial Neural Networks, and Bayesian analysis may be applied to model the tessellation data, to check their effectiveness. Extensive parameter tuning of K-Means should be done, if possible, to improve the clustering. K-Means algorithm can also be replaced by density based clustering algorithms; for example, DBSCAN and OPTICS, among others.
 
 \section*{Acknowledgements}
 The work is undertaken as part of the Cognizant-Ernet IIT Madras project. The authors are also thankful to Vishnu Raj, Gopal Krishna Kamath, and Sudharsan Parthasarathy for many helpful discussions.


\end{document}